\documentclass[sigconf]{acmart}

\AtBeginDocument{%
  }

\setcopyright{acmlicensed}
\copyrightyear{2018}
\acmYear{2018}
\acmDOI{XXXXXXX.XXXXXXX}
\acmConference[Conference acronym 'XX]{Make sure to enter the correct
  conference title from your rights confirmation email}{June 03--05,
  2018}{Woodstock, NY}
\acmISBN{978-1-4503-XXXX-X/2018/06}

\usepackage{booktabs}
\usepackage{longtable}
\usepackage{array}
\usepackage{multirow}
\usepackage{graphicx}

\newcommand\todo[1]{\textcolor{blue}{#1}}

\usepackage{soul} 

\begin{document}

\title{Beyond Effectiveness: A Multi-Criteria Framework for Comparing Practical Socio-Technical Interventions}

\author{Catherine King}
\affiliation{%
  \institution{Carnegie Mellon University}
  \city{Pittsburgh}
  \country{USA}}
\email{cking2@andrew.cmu.edu }

\author{Lynnette Hui Xian Ng}
\affiliation{%
  \institution{Carnegie Mellon University}
  \city{Pittsburgh}
  \country{USA}
}
\email{lynnetteng@cmu.edu}

\author{Kathleen M. Carley}
\affiliation{%
 \institution{Carnegie Mellon University}
 \city{Pittsburgh}
 \country{USA}}
 \email{kathleen.carley@cs.cmu.edu}

\renewcommand{\shortauthors}{Anonymous}

\begin{abstract}
Designers and policymakers in sociotechnical domains like content moderation, privacy interfaces, recommender systems and beyond, must choose among a growing menu of proposed interventions, but typically lack a principled basis for comparing them. Prior work tends to evaluate interventions individually and mostly along the effectiveness criteria, while implementation constraints such as cost, effort and feasibility are often considered separately. We present a multi-criteria framework for evaluating sociotechnical interventions. This framework is instantiated through the case of misinformation, a domain of intense focus for proposed countermeasures. We survey $N=39$ researchers on 40 operationalized interventions across five evaluative criteria: political feasibility, effectiveness, user acceptance, cost, and implementation effort. We find that the interventions that experts judge to be the most effective are not always the most acceptable to the public or the most feasible to implement. We also discuss how this tension has implications for the design of sociotechnical interventions beyond misinformation, and offer a decision framework for practitioners navigating the trade-offs of sociotechnical interventions.
\end{abstract}



\received{20 February 2007}
\received[revised]{12 March 2009}
\received[accepted]{5 June 2009}

\maketitle

\section{Introduction}
Social media platforms and the emerging generative-AI technology that shapes them are quickly transforming the online information environment, bringing with them various interventions used to address socio-technical harms~\citep{kingPublicSupportMisinformation2026,schoenebeck2023online,carley2020social}. Yet these interventions, such as those for countering misinformation or detecting fraud, are often developed and treated individually rather than analyzed as a whole using a standardized set of criteria \cite{courchesneReviewSocialScience2021,greenEvidenceBasedMisinformationInterventions2022}. Socio-technical systems research has argued that such interventions cannot be meaningfully evaluated outside the social context in which they enter~\citep{selbst2019fairness}, and value-sensitive design calls for weighing a technology against multiple, sometimes competing values rather than a single metric~\citep{friedman2019value}. Drawing on these two threads of research, we aim to establish a comprehensive framework for developing effective and practical countermeasures across the social media landscape. Researchers, companies, and policymakers can use this framework to evaluate the socio-technical interventions and consider their trade-offs.

We develop a multi-criteria framework for socio-technical intervention design through the case of misinformation. 
Throughout, we use "intervention" to mean a deliberate countermeasure (i.e., a policy, a platform design, an educational effort) that is deployed to reduce a socio-technical harm.
Misinformation is a domain where the evaluation gap is especially visible. Countermeasures are proliferating faster than the field's capacity to compare them ~\citep{varela2025countermeasures,reuter2025combating}, and platforms tend to implement interventions reactively and inconsistently~\citep{gillespie2018custodians}. Intervention design decisions typically require trading effectiveness against the practical and normative considerations that determine whether an intervention can or should be be deployed~\citep{grelleWhenWhyPeople2024,rochefortRegulatingSocialMedia2020}. However, to the best of our knowledge, no framework exists to support that reasoning. To develop the framework, we integrate prior research with an expert elicitation survey of $N=39$ researchers who evaluated 40 interventions across five criteria: political feasibility, effectiveness, user acceptance, cost, and effort level. Our main \textbf{research question} is: How can we identify which interventions are both practical and effective, and under what circumstances? More specifically:
\begin{itemize}
    \item RQ1: What trade-offs emerge when misinformation interventions are evaluated across political feasibility, effectiveness, acceptance, cost, and effort?
    \item RQ2: Which intervention categories best balance user-centered and implementation-centered criteria?
    \item RQ3: What design principles can guide practical misinformation intervention design?
\end{itemize}

The contributions of this work are threefold: 
\begin{itemize}
    \item First, we introduce a taxonomy of forty operationalized interventions based off a harmonization of literature review. These interventions are structured around eight categories, consisting of account moderation, content moderation, content distribution, generative AI-specific measures, content labeling, user-based measures, media literacy and educational efforts, and other external and institutional measures.
    \item Second, we provide a multi-criteria expert evaluation of those interventions across five criteria: political feasibility, effectiveness, user acceptance, cost, and implementation effort.  This expert elicitation survey shows that interventions frequently vary in effectiveness, acceptance or political feasibility, and the administrative effort or cost involved.
    \item Third, we translate these judgments into a decision framework that maps practical operating constraints (budget, political exposure, timeline, public trust) onto a starting cluster of interventions, and the design investments needed to unlock the next cluster.
\end{itemize}

Designers and policymakers often face similar tensions across many sociotechnical domains: how to moderate content~\citep{jiangTradeoffcenteredFrameworkContent2023}, how much friction to add to privacy choices~\citep{acquisti2017nudges}, or how to regulate dark patterns in interface design~\citep{mathur2019dark}. In each case, the question is not whether an intervention works, but which one to deploy given the contextual constraints. Here, we study misinformation as a case study to build a tool for this selection problem. This study provides insights that we discuss on how and which intervention types, scoped to the social media information environment, are most viable across different implementation constraints, and discusses how different stakeholders can use this framework. 

\section{Related Work}

\subsection{Misinformation interventions and their evaluation}
Previous systematic review articles in the domain of misinformation interventions show that interventions are typically characterized by the part of the platform affected or the actor implementing them. \citet{blairInterventionsCounterMisinformation2024} groups interventions into four general categories (informational, educational, sociopsychological, and institutional) and evaluates them primarily on efficacy, while suggesting that other dimensions like feasibility, scalability, and durability should also be considered when selecting and designing interventions. \citet{courchesneReviewSocialScience2021} find a mismatch between the countermeasures that are studied versus those that are implemented by platforms. Specifically, fact-checking interventions and interventions emphasizing source pre-bunking or user inoculation are widely studied, while account-level interventions such as removal and user notifications, and algorithmic interventions such as redirection and changes to monetization policy, remain understudied. Finally \citet{kozyrevaToolboxIndividuallevelInterventions2024} categorizes interventions into three general categories (nudges, boosts and educational interventions, and refutation strategies) and organizes implementation through the targeted outcome (i.e. behaviors, competence, and beliefs) and limitations. 

While these three key review articles anchor the space of misinformation interventions, they stop short of the comparative evaluation that design practice needs. These review articles find that most prior research in this domain focuses narrowly on effectiveness or user acceptance of the interventions, but rarely both axes, and almost never alongside implementation constraints. Our framework closes this gap by comparing multiple dimensions of interventions at once to facilitate identifying the appropriate interventions for the context.

\subsection{Intervention design features and trade-offs}
To design and evaluate interventions based on their features using multiple criteria, we first need to define what those features are. A well-defined intervention has a set of descriptive, objective characteristics, as well as potentially changing user perceptions of those characteristics and the intervention as a whole. 
An intervention’s objective characteristics refer to the relatively static features that define how it is implemented and by whom. These features include the implementer (which organization(s) are implementing the intervention), the targeted phase of the information pipeline (such as the creation, spread, or belief phases), the content being addressed, platform design changes, and the policy’s description (how it is being implemented and communicated)~\citep{king2025thesis}.
User perception refers to how individuals evaluate the different objective characteristics of interventions and includes factors such as perceived effectiveness, fairness, intrusiveness, transparency, and problem awareness ~\citep{grelleWhenWhyPeople2024}. Interventions that are viewed as more effective (benefits outweigh costs) and are more supported by users (practical) are more likely to be implemented by platforms \citep{gorwaWhatPlatformGovernance2019,kingPublicSupportMisinformation2026,liu2022implications}. 

User perceptions are especially critical because they influence the overall level of policy support. These perceptions affect whether people believe a problem even needs intervention, whether they think the intervention will be effective and worth doing, and whether they trust the proposed implementer to enforce the new policies fairly. Public policy support often depends on problem awareness and the type of content addressed~\citep{grelleWhenWhyPeople2024}.
There is a higher level of support for anti-smoking initiatives compared to alcohol consumption or diet, which is partially explained by the public’s awareness of the negative consequences associated with smoking~\citep{diepeveenPublicAcceptabilityGovernment2013}. Similarly, in a social media context, individuals have indicated they are more willing to engage in social corrections with close contacts or for seriously harmful content~\citep{king2025brims,tandocDiffusionDisinformationHow2020,king2025path}.
Additionally, some individuals may trust certain implementers more than others. For example, partisanship is associated with lower levels of support for government regulation~\citep{vogelsSupportMoreRegulation2022}.

Educational and messaging efforts could improve public perceptions and support for interventions. A meta-analysis of public policy interventions demonstrated that effectively communicating a potential policy's effectiveness increased overall support by approximately 4\% ~\citep{reynoldsCommunicatingEffectivenessIneffectiveness2020}. 
Interventions that are more transparently implemented also tend to gather more public support ~\citep{grelleWhenWhyPeople2024,jiangTradeoffcenteredFrameworkContent2023}. Clearly marked and communicated interventions can shape a user's perception of the effectiveness and intrusiveness of the intervention~\citep{zhouEffectiveReportingSystem2024}. However, more opaque policies may in turn be less exploitable by malicious actors~\citep{jiangTradeoffcenteredFrameworkContent2023}, illustrating the trade off with transparency.

\subsection{Criteria selection for platform design}
The field of Human-Computer Interaction has a substantial body of work on how platform affordances and intervention design shape user information behavior. Most of this work has independently arrived at the same trade-off-centered framing, where there are multiple criteria to balance simultaneously. Content moderation work treats moderation not as a single optimization target but as a negotiation among competing values of efficiency, fairness, and user trust~\citep{gillespie2018custodians}. Ethnographic accounts of Reddit's moderation tool describe this negotiation as part of the platform's broader sociotechnical system, where automated and human judgment must be jointly accounted for~\citep{wright2022automated}. Comparable trade-offs structure privacy interface design, where the amount of friction a nudge introduces is weighed against its protective benefit rather than maximized outright~\citep{acquisti2017nudges}. Within misinformation specifically, recent reviews converge on effectiveness, feasibility, scalability, and durability as the relevant criteria~\citep{blairInterventionsCounterMisinformation2024,kozyrevaToolboxIndividuallevelInterventions2024}. Feasibility refers to how easily an intervention can be implemented and whether it aligns with user expectations and the intervention’s intent~\citep{greenEvidenceBasedMisinformationInterventions2022}. Scalability and durability, both aspects of effectiveness, address how easily an intervention can be expanded to reach a larger audience and the longevity of its impact, respectively ~\citep{huangMediaLiteracyInterventions2024,maertensLongtermEffectivenessInoculation2021,porterGlobalEffectivenessFactchecking2021}. 

These criteria focus on the technical components or user interactions with platforms, without considering external influences outside the platforms. For that, we turn to public policy research. In public policy, when comparing strategies for regulating socio-technical platforms, research often evaluates proposals using measures like effectiveness, administrative difficulty or burden, cost, and political acceptability \cite{rochefortRegulatingSocialMedia2020,kraftPublicPolicyPolitics2017,dudleyRegulationPrimer2012}. The public policy perspective on administrative burden relates to the concept of implementation feasibility. Additionally, political acceptability is a similar concept to user acceptance of an intervention, which has been shown to influence support for that intervention, though it also incorporates external governmental constraints~\citep{rochefortRegulatingSocialMedia2020,kingPublicSupportMisinformation2026,gorwaWhatPlatformGovernance2019}.

\section{The PEACE Framework}
In this work, we first define a categorization of intervention types and then develop a decision framework for evaluating intervention quality and balancing trade-offs. 
The \textbf{PEACE} framework assesses each intervention’s \textbf{Political feasibility} in the current environment, its \textbf{Effectiveness} in achieving the intended outcome, the level of user \textbf{Acceptance} it is likely to receive, its \textbf{Cost} or cheapness, and the \textbf{Effort} required for implementation. These five evaluative criteria were drawn from literature, and insights from an expert elicitation survey was used to inform the design of a corresponding decision framework. 

\subsection{Intervention Categorization}
To categorize intervention types, we selected and examined all the studies cited by four recent, comprehensive review articles from different fields: the \citet{courchesneReviewSocialScience2021} article published in an interdisciplinary journal focused on social sciences, the \citet{blairInterventionsCounterMisinformation2024} article highlighting interventions from both the Global North and Global South in a psychology journal, the \citet{aghajariReviewingInterventionsAddress2023} article published in the proceedings of a prominent HCI conference, and the \citet{kozyrevaToolboxIndividuallevelInterventions2024} article in Nature Human Behavior. We augmented the literature review by pulling additional, relevant articles on socio-technical interventions identified through a Scopus search using the terms ``misinformation'' and ``disinformation'' combined with either ``interventions'' or ``counter''. We then categorized interventions on social media platforms by the stages of the information life cycle they target \citep{king2025icwsm}.
More details on our categorization and the specific literature referenced can be found in \autoref{sec:app_literature_interventions}.

The information lifecycle on social media generally includes three phases: the creation of information-disseminating accounts and the information itself, the dissemination of information through organic and viral amplification, and the engagement with the information by correcting, contesting, or contextualizing it~\citep{kingPublicSupportMisinformation2026,ng2021does,wardleInformationDisorderInterdisciplinary2017,ciampagliaDigitalMisinformationPipeline2018}.
The life cycle lens provides a frame for thinking about where analysts could intervene. Because the volume and speed of information spread are too fast for journalists and users to respond promptly to incorrect or contested claims, such as with fact-checks, exploring ways to intervene earlier in the pipeline, like slowing the virality of misinformation, is an important consideration~\citep{ciampagliaDigitalMisinformationPipeline2018}. Table \ref{tab1} provides a high-level overview of the intervention categories that will be discussed throughout the rest of the paper \cite{kingPublicSupportMisinformation2026}. Refer to Appendix~\ref{sec:app_detailed_interventions} for detailed information on the 40 specified intervention implementations. 
The appendix outlines how governments, platforms, and institutions can implement these measures in practice. The specific interventions were operationalized by reviewing the literature and platform policies to identify examples for each intervention type that have been implemented or proposed previously \cite{DemocracyDesignSocial2024,rochefortRegulatingSocialMedia2020}.

\begin{figure*}[htp]
\centering
\includegraphics[width=.8\textwidth]{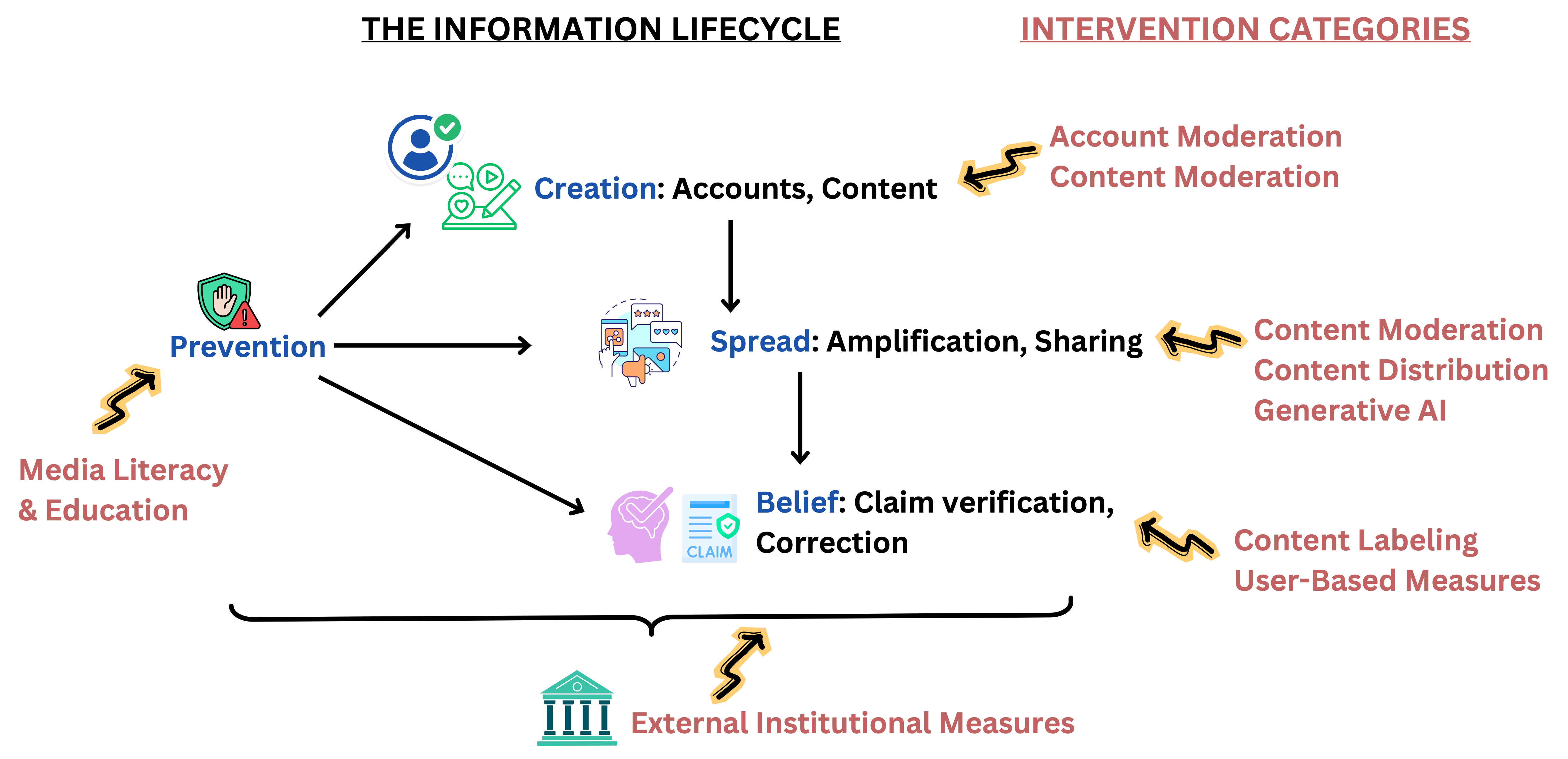}
\caption{The information lifecycle~\citep{kingPublicSupportMisinformation2026} and corresponding intervention categories. } \label{fig:information_lifecycle}
\Description[The information lifecycle and corresponding intervention categories.]{The creation step is associated with account and content moderation. The spread step is associated with content moderation, distribution, and generative AI-specific interventions. The belief step is associated with content labeling and user-based measures. Prevention can affect all the three steps and includes media literacy and education. Institutional measures are external to the lifecycle.}
\end{figure*} 

\begin{table*}[htp]
\centering
\caption{Categories of Socio-Technical Interventions, based on the information lifecycle~\citep{kingPublicSupportMisinformation2026}}\label{tab1}
\begin{tabular}{>{\raggedright\arraybackslash}p{3.7cm}>{\raggedright\arraybackslash}p{2.9cm}>{\raggedright\arraybackslash}p{7.4cm}}
\toprule
Information Lifecycle & Intervention Category & Definition \\ 
\midrule
\textbf{Creation}: accounts, content & Account Moderation & The moderation of user accounts such as by suspending, banning, or limiting users \\ 
\textbf{Spread}: viral amplification, individual sharing & Content Moderation & The moderation of content such as by removing, downranking, or correcting it \\ 
& Content Distribution & Affecting the distribution/ sharing of content such as by using redirection, accuracy prompts, or friction\\ 
 & Generative AI-specific & Employing genAI to address harmful or low-quality content (e.g., using chatbots to dialogue with users, using genAI to produce fact checks)\\
\textbf{Belief}: claim verification, correction & Content Labeling & The use of labeling or disclosure to notify users, provide additional context, or verify claims \\
& User-based Measures & Relating to user-driven responses to the information, such as social corrections or reporting \\
\textbf{Prevention} & Media Literacy and Education & Training efforts aimed at improving the public's media literacy and critical thinking skills \\ 
\midrule
External & Institutional Measures & Measures taken outside the platform ecosystems such as regulation, data sharing, and investing in journalism \\
\bottomrule
\end{tabular}
\end{table*}

The first six categories (account moderation, content moderation, content distribution, generative AI-specific, content labeling, user-based measures) are interventions that occur directly on the platforms. These interventions can be implemented by social media companies, users on those platforms, or imposed on the platforms through government regulation~\citep{rochefortRegulatingSocialMedia2020}. These platform-level interventions fall into two broad mechanisms: algorithmic interventions that govern what accounts and content can be created and distributed (account moderation, content moderation, content distribution, generative AI-specific) and front-end design choices that govern how content is displayed and the affordances in which it can be engaged with (content labeling, user-based measures). More specifically, account moderation strategies target the information creation phase by controlling account and content creation permissions; content moderation, distribution, and generative AI-assisted strategies target the information spread phase by reducing or removing algorithmic amplification and managing the information sharing process between users; and content labeling and user-driven strategies occur after the information has been shared to verify or contextualize claims. 

The next two categories, media literacy and external institutional measures, are strategies that occur primarily beyond the platform, and can be implemented by platforms, governments or various civic organizations. Media literacy and educational initiatives focus on the prevention phase in the information pipeline, teaching students how to identify and discern truth from misinformation and scams. This includes learning how to recognize manipulation techniques, logical fallacies, and AI-generated or edited content. Structural strategies like regulation and data sharing may target any part of the pipeline, and are meant to support a healthier information environment.

In this study, we specifically examine the case of misinformation spread through social media platforms. The information lifecycle includes the creation of social media accounts and misinformation, the dissemination of misinformation, and the correction or prevention of false beliefs and actions. The same lifecycle lens applies to adjacent harms -- scams, fraud, coordinated inauthentic behavior, terrorist recruitment etc-- that move through creation, spread, and belief on the same platforms. 

\subsection{PEACE Criteria}

For this study, five evaluative criteria were selected after reviewing previous comprehensive review articles and policy analyses~\citep{blairInterventionsCounterMisinformation2024,kozyrevaToolboxIndividuallevelInterventions2024,rochefortRegulatingSocialMedia2020}. These criteria are: \textbf{political feasibility}, \textbf{effectiveness}, \textbf{acceptance}, \textbf{cost}, and \textbf{effort level}. The five main criteria are defined as follows:

\begin{itemize}
     \item \textbf{Political feasibility:} The likelihood of implementing an intervention, incorporating the following factors as well as external, political factors, such as stakeholder perspectives, regulatory and legal constraints, and support from relevant officials or organizations. 
     \item \textbf{Effectiveness:}  The extent to which an intervention successfully targets the creation, spread, or belief phases in the information lifecycle under different circumstances, such as for some users, certain platforms, or particular types of content.
     \item \textbf{Acceptance:} The degree of user support for an intervention, which may be influenced by factors such as its perceived effectiveness, fairness, intrusiveness, transparency, and the level of trust in the implementer. 
     \item \textbf{Cost:} The financial burden on the users, platforms, or government entities involved in implementing and maintaining the intervention. 
     \item \textbf{Effort:} The level of administrative effort required to implement and maintain the intervention for the users, platforms, or government entities involved.  
\end{itemize}

The PEACE decision framework provides a balanced overview of each intervention's overall viability. The inherent characteristics of an intervention, along with how users perceive it, directly affect these five key evaluation criteria. More specifically, user perceptions can impact the acceptance of the intervention, thereby affecting public support for the policy, which in turn increases policy compliance and effectiveness~\citep{grelleWhenWhyPeople2024}. For example, political feasibility likely affected Meta's US fact-checking program in 2025, where despite its effectiveness and broad user acceptance~\citep{randAmericansActuallyWant2025}, external political pressures led to the program's termination~\citep{zhanHeresWhyMeta2025,kaplanMoreSpeechFewer2025}.

To provide a comprehensive set of recommendations across the intervention landscape that researchers, companies, and policymakers can use, we conducted an expert elicitation survey based on these five evaluative criteria. 
Design and decision research have long relied on structured expert judgment when the outcome data for a given intervention is costly, slow or impossible to obtain at the scale of comparison against alternatives~\citep{o2006uncertain}. Within misinformation research specifically, survey-based expert elicitation has previously been used to gauge which interventions experts believe should or would be effective, providing direct precedent for soliciting comparative judgments across our criteria, rather than waiting for outcome evidence for multiple interventions individually~\citep{altaySurveyExpertViews2023,blairInterventionsCounterMisinformation2024}. 

\section{Data and Methods} 

\subsection{Survey Overview}
We conducted an expert elicitation survey to obtain comparative judgments of across the five evaluative criteria of political feasibility, effectiveness, user acceptance, cost, and effort level. 
The survey was promoted to participants and invitees of the \textbf{SUMMIT 2025} \textbf{[Full summit name, host location, and dates removed for peer review to preserve anonymity.]} Our use of expert elicitation for survey judgment captures how experts anticipate the impact of interventions, including their effectiveness in combatting misinformation, how the public and users are likely to respond to the interventions, and their expert evaluation of each intervention's feasibility, cost, and implementation effort.


The survey data was collected between January 23 and March 24, 2025 from thirty-nine researchers. The experts provided their professional assessments of the effectiveness and acceptability of the general intervention categories (\autoref{tab1}) using a 1-7 Likert scale. They also each evaluated 12 out of the 40 operationalized interventions (\autoref{tab:operationalized-interventions}) across the five evaluative metrics using a 1-5 Likert scale. 
 The specific interventions assigned to each participant were randomly selected. Because of the survey’s length, participants were not assigned all 40 interventions. Instead, 12 were assigned to each participant to ensure about 10 (or more) responses per intervention. Informed consent was obtained from all participants. See the supplemental file for a complete copy of the survey.

In the survey, a "1" on the Likert scale indicated low political feasibility, effectiveness, acceptance, cost, and effort; and "5" indicated high. During the analysis, cost and effort level scores were inverted (i.e., a Likert score of "5" indicated low cost and effort), ensuring all five metrics ran in the same direction, with higher scores reflecting better results. The metrics were also renamed to \textbf{cheapness} and \textbf{ease} of implementation for clarity in the Results section. This made it possible to analyze an overall average score for each intervention and conduct a cluster analysis.

\subsection{Demographics}
The demographics of the participants who completed the survey are shown in Table~\ref{tab:demographics}.

\begin{table}[htbp]
\centering
\small
\caption{Participant Demographics (N = 39)}
\label{tab:demographics}
\begin{tabular}{llrr}
\toprule
\textbf{Category} & \textbf{Response} & \textbf{N} & \textbf{\%} \\
\midrule
\multirow{3}{*}{Gender}
    & Women                     & 16 & 41.0 \\
    & Men                       & 22 & 56.4 \\
    & Other / prefer not to say & 1  & 2.6  \\
\midrule
\multirow{4}{*}{Age}
    & 18--34                    & 15 & 38.5 \\
    & 35--44                    & 12 & 30.8 \\
    & 45+                       & 10 & 25.6 \\
    & Prefer not to say         & 2  & 5.1  \\
\midrule
\multirow{5}{*}{Academic Rank}
    & Faculty                   & 21 & 53.8 \\
    & PhD Student                & 12 & 30.8 \\
    & Industry                  & 2  & 5.1  \\
    & Postdoc                   & 1  & 2.6  \\
    & Other                     & 3  & 7.7  \\
\midrule
\multirow{6}{*}{Primary Discipline}
    & Computational Social Sciences        & 17 & 43.6 \\
    & Computer Science                     & 10 & 25.6 \\
    & Sociology                            & 4  & 10.3 \\
    & Communications and Media Studies     & 4  & 10.3 \\
    & Political Science                    & 2  & 5.1  \\
    & Other                                & 2  & 5.1  \\
\bottomrule
\end{tabular}
\end{table}

\section{Results and Analysis}

\subsection{Category-Level Tradeoffs}
\begin{figure*}[htp]
\centering
\includegraphics[width=\textwidth]{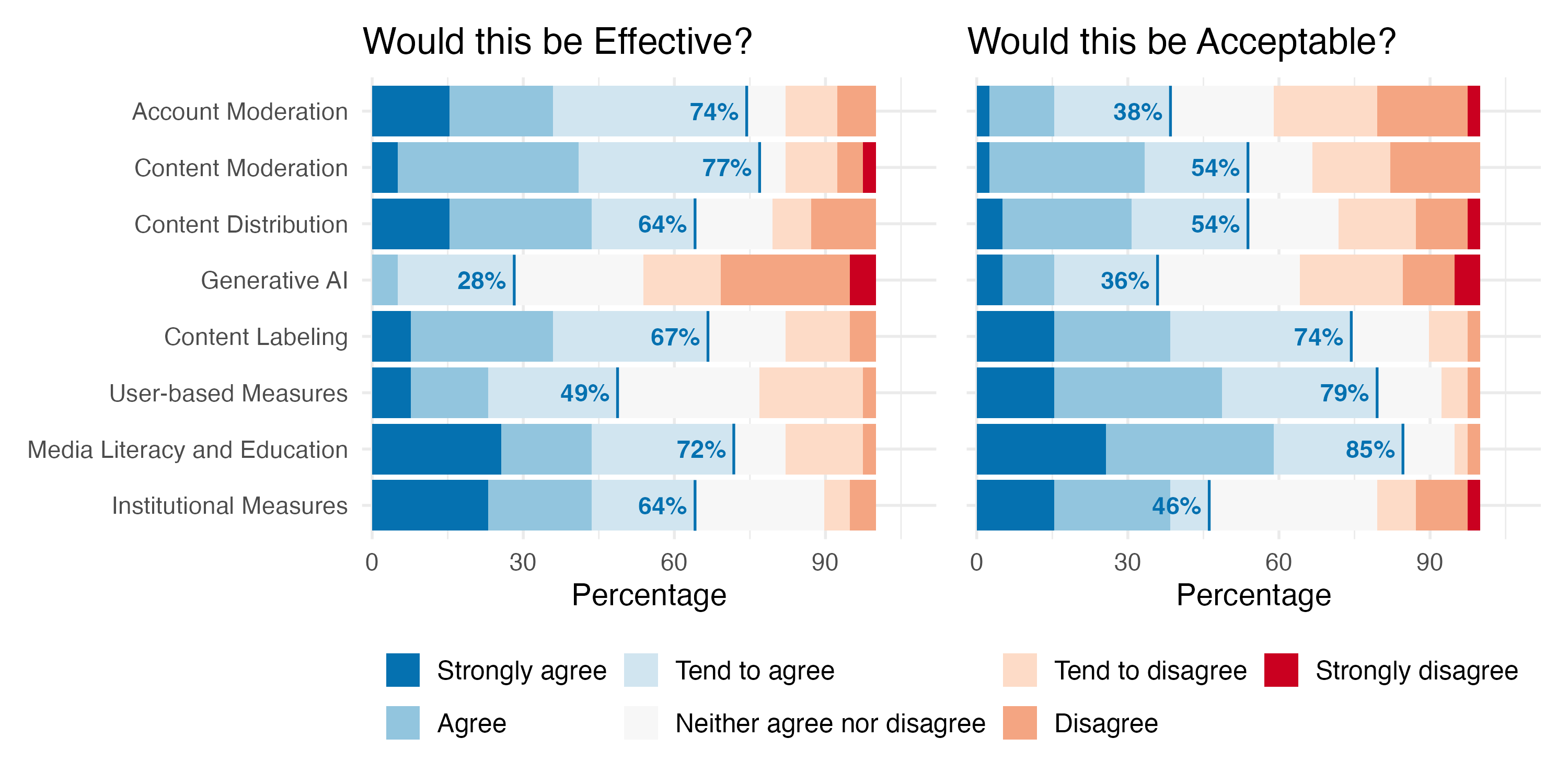}
\caption{Stacked bar plot showing the percentage of responses on the effectiveness and acceptance of the general intervention categories.} \label{fig:general_eff_acc}
\Description[Stacked bar plot showing the percentage of responses on the effectiveness and acceptance of the general intervention categories.]{The sub-figure on the left shows the responses for effectiveness, with the percentage of experts who at least somewhat agree that category of interventions would be effective highlighted. Content moderation, account moderation, and media literacy score above 70\%, content labeling and distribution above 60\%, user-based measures at around 50\% and genAI at 28\%. The sub-figure on the right shows the responses for acceptance, with the percentage of experts who at least somewhat agree that category of interventions would be acceptable to users highlighted. Media literacy, user-based measures, and content labeling are above 70\%. Content moderation, distribution, and institutional measures are around 50\% , and account moderation and genAI are just below 40\%. }
\end{figure*}

To better understand how experts perceive the misinformation interventions landscape, we first asked participants to rate the overall effectiveness and user acceptance of each of the 8 intervention categories among the American public. Figure \ref{fig:general_eff_acc} shows the fraction of participants who believed each category of countermeasures to be effective or acceptable to the public. Content moderation, account moderation, and media literacy / education are regarded as the most effective, with over 70\% of respondents agreeing they would be effective. However, moderation, especially account moderation, is viewed by experts as one of the least acceptable to the public.

\begin{figure*}[htp]
\centering
\includegraphics[width=.8\textwidth]{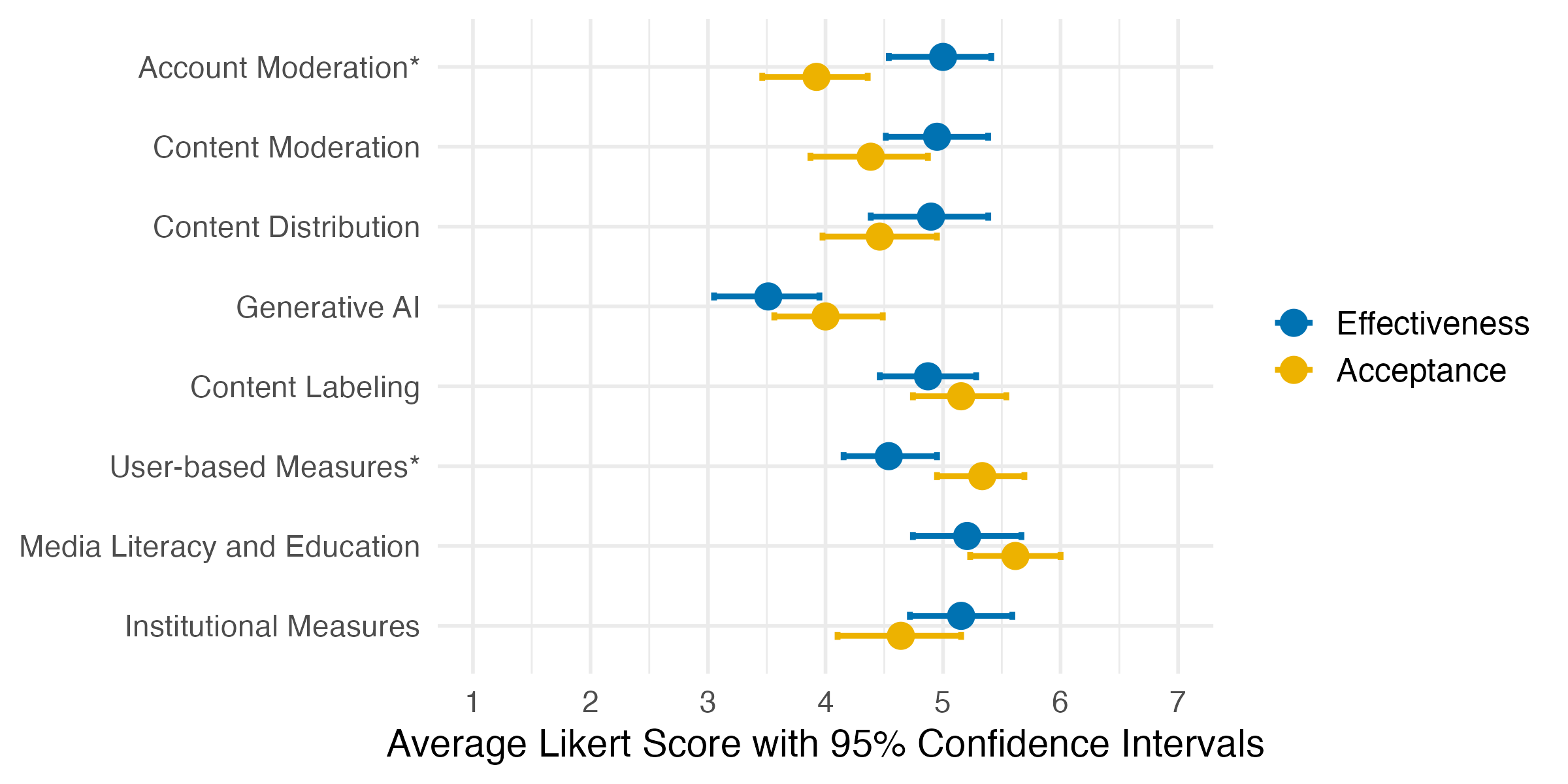}
\caption{Average effectiveness and acceptance Likert scores per general intervention category.} \label{fig:general_error_plot}
\Description[Average effectiveness and acceptance Likert scores per general intervention category.]{Average effectiveness and acceptance Likert scores per general intervention category. Account moderation and user-based measures are the only categories where the 95\% confidence intervals of the average effectiveness and acceptance Likert scores do not overlap, with account moderation having higher effectiveness scores and user-based measures having higher acceptance scores. }
\end{figure*}

To further investigate the apparent tension between effectiveness and acceptance, Figure \ref{fig:general_error_plot} illustrates the average Likert scores and 95\% confidence intervals for the effectiveness and acceptance ratings. In this figure, we observe that media literacy / education and external institutional measures rise in the effectiveness rankings, bolstered by the proportion of participants who ``strongly agree" regarding their effectiveness rather than more weakly agreeing. Content and account moderation remain at high levels. While Figure \ref{fig:general_eff_acc} shows general agreement or disagreement among experts, this figure incorporates the strength of their opinions by using the Likert scores directly, making the differences in effectiveness and acceptance across some intervention categories more apparent. Account moderation and user-based measures exhibit the largest gap between effectiveness and acceptance, with effectiveness exceeding acceptance in account moderation, and acceptance exceeding effectiveness in user-based measures (both statistically significant at $p < 0.05$).

These results align with a similar survey of experts, conducted by ~\citet{altaySurveyExpertViews2023}, which found that experts believed that media literacy, labeling of false content, and fact-checking are the most effective interventions. Each of these interventions received around 70\% approval. Meanwhile, accuracy prompts (a form of content distribution), source labeling, and inoculation ranked lower on the list. These were the only interventions included in that survey in a similarly structured question~\citep{altaySurveyExpertViews2023}. \citet{altaySurveyExpertViews2023} also asked experts if certain types of actions \textit{should} be taken, and concluded that platform design changes, algorithmic changes, and general content moderation as the most popular (over 75\% agree), and penalizing misinformation sharing and shadow banning (a form of account moderation) as the least popular. In fact, shadow banning was the only suggested intervention where more experts disagreed with its implementation than agreed. Similarly,~\citet{blairInterventionsCounterMisinformation2024} found in a survey of experts that they recommended implementing educational and institutional interventions (such as media literacy, platform alterations, and journalist training) over other types of interventions. However, the authors note that these interventions tend to be among the least well-studied or have mixed evidence on actual effectiveness \cite{blairInterventionsCounterMisinformation2024}. 

\subsection{Metric Tradeoffs}

\begin{figure*}[htp]
\centering
\includegraphics[width=\textwidth]{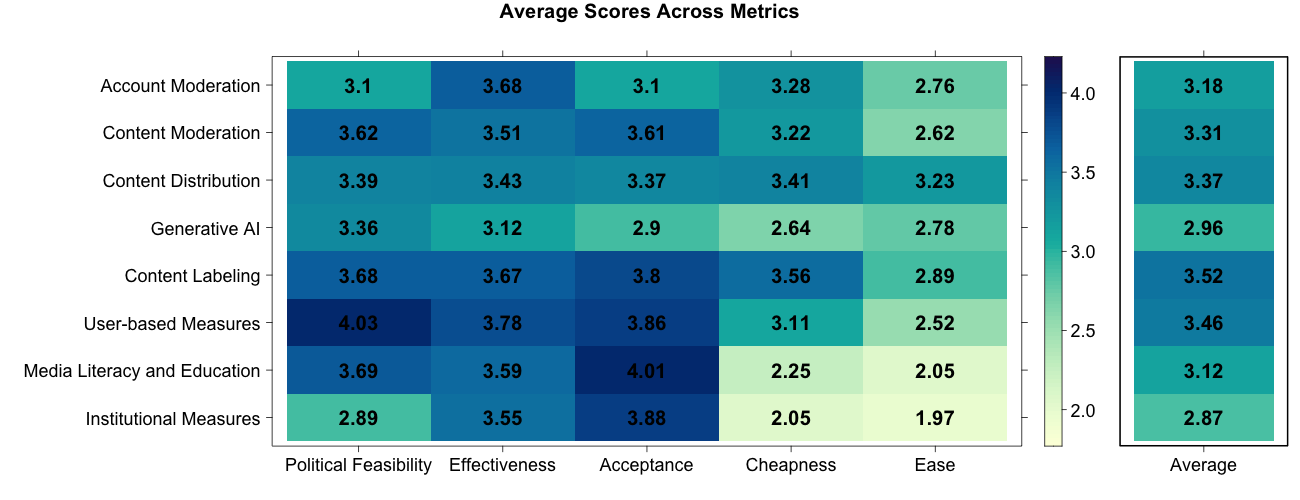}
\caption{A heatmap showing the average score for each metric across general intervention categories.} \label{fig:heatmap}
\Description[A heatmap showing the average score for each metric across general intervention categories.]{A heatmap showing the average score for each metric across general intervention categories. Most categories perform better on certain criteria than others. The most well-balanced category, and the only one that averages above a 3 on the Likert scale for each criterion, is content distribution. }
\end{figure*}

Beyond effectiveness and acceptance of the high-level intervention categories, we also asked the experts to rate each individual intervention across five criteria: political feasibility, effectiveness, user acceptance, cheapness (inverted cost score), ease of implementation (inverted effort score). 
Figure \ref{fig:heatmap} presents a heatmap that illustrates the overall average metric scores, averaged by category and ranked from highest to lowest. 
Content labeling, user-based measures, and content distribution topped the list among the five criteria. Content labeling and user-based measures scored well on effectiveness, acceptance, and feasibility. Content distribution was the most balanced category with the most similar average metric values and the only category where all five metrics averaged above a 3 on the Likert scale. Media literacy and external institutional measures were ranked last, primarily due to their high overall costs (low affordability) and required effort level. Taken together, these results suggest that while some countermeasures are seen as effective, and some are even considered both effective and acceptable by the public, they are not always regarded as practical by experts. Interventions that require more administrative effort or resources to implement tend to rank lower overall, despite their potential positive impact.

\subsection{Top Interventions}
\label{sec:top_interventions}
To identify which individual interventions drive these category-level tradeoffs, and what features they may have in common, we rank interventions by their overall scores (average of the five metrics) in Table \ref{tab:intervention_rankings}. In general, interventions in the content labeling, user-based measures, and content distribution interventions dominate the top 10. Only two interventions from moderation categories (user control and demonetization) rank in the top 10 overall, and these moderation interventions are among the least restrictive or provide users with a high level of agency compared to other moderation interventions. 

\begin{table*}[htp]
\centering
\small
\caption{Interventions ranked by adjusted average score.  The top-ranked intervention in each category is bolded.}
\begin{tabular}{p{3cm} p{2.7cm} p{.65cm} | p{3.2cm} p{2.8cm} p{.65cm}}
\toprule
\textbf{Intervention Rank} & \textbf{Category} & \textbf{Score} & \textbf{Intervention Rank} & \textbf{Category} & \textbf{Score} \\
\midrule
\textbf{1. Government Labels    }  & \textbf{Content Labeling}   & \textbf{4.04} & 21. Virality Circuit Breakers & Content Moderation  & 3.23 \\
\textbf{2. Limit Resharing  }      & \textbf{Content Distribution }  & \textbf{3.91} & 22. Ban Deepfakes                 & Content Moderation  & 3.21 \\
3. Limit Forwarding       & Content Distribution   & 3.91 & 23. Platform Alterations      & Content Distribution   & 3.20 \\
4. Crowdsourcing         & Content Labeling    & 3.76 & \textbf{24. Data Sharing}              & \textbf{Institutional Measures } & \textbf{3.20} \\
5. AI Disclosure          & Content Labeling  & 3.62 & 25. Warning Labels & Content Labeling  & 3.18 \\
6. Friction               & Content Distribution   & 3.55 & 26. Redirection               & Content Distribution    & 3.16 \\
\textbf{7. Reporting   }           & \textbf{User-based Measures }  & \textbf{3.52} & \textbf{27. Digital Media Literacy  }  & \textbf{Media Literacy }  & \textbf{3.13} \\
\textbf{8. User Control   }        & \textbf{Content Moderation}  & \textbf{3.45} & 28. Inoculation               & Media Literacy  & 3.10 \\
9. Alter Platform Metrics & User-based Measures  & 3.45 & 29. Ban Political Ads         & Content Distribution   & 3.06 \\
\textbf{10. Demonetization }       & \textbf{Account Moderation} & \textbf{3.43} & 30. Account Suspensions       & Account Moderation & 3.01 \\
11. Social Norms          & User-based Measures   & 3.41 & 31. Deplatforming             & Account Moderation & 2.98 \\
12. Targeted Ads          & Institutional Measures  & 3.41 & 32. \textbf{Gen AI Chatbots }          & \textbf{Generative AI}   & \textbf{2.96} \\
13. Algorithmic Changes   & Content Moderation   & 3.36 & 33. Gen AI Content            & Generative AI   & 2.96 \\
14. Accuracy Prompts      & Content Distribution   & 3.35 & 34. Fact-Check Ads            & Content Distribution    & 2.91 \\
15. Content Removal       & Content Moderation   & 3.31 & 35. Journalism Support        & Institutional Measures  & 2.89 \\
16. Downranking           & Content Moderation   & 3.31 & 36. Media Support             & Institutional Measures  & 2.86 \\
17. Shadowbanning        & Account Moderation & 3.31 & 37. Taxes / Fines             & Institutional  Measures  & 2.85 \\
18. Source Labeling       & Content Labeling   & 3.27 & 38. Gov't Regulation     & Institutional Measures  & 2.69 \\
19. AI in Ads             & Content Distribution  & 3.26 & 39. Anti-Trust Action         & Institutional Measures  & 2.62 \\
20. Fact-check Labels     & Content Labeling   & 3.24 & 40. Privacy Legislation       & Institutional Measures  & 2.45 \\
\bottomrule
\end{tabular}
\label{tab:intervention_rankings}
\end{table*}

Analyzing only the highest overall scores among individual interventions provides a limited view and fails to highlight the trade-offs among the different metrics. When diving deeper into the top ten interventions, we find that only three are also in the top ten for effectiveness, five for acceptance, effort, and cost, and seven for political feasibility; the interventions rate differently for different metrics. Overall, most of the top interventions rank in the top ten for only two or three metrics. To better analyze the trade-offs, we used base R's implementation of the Ward's D2 clustering method to group interventions based on similar criterion scores \citep{murtagh2014ward,baseR2025}. We identified four clusters that summarize recurring trade-off patterns. Table \ref{tab:ward-clusters} presents the average criteria values for each intervention in each cluster:
\begin{enumerate}
        \item \textit{Balanced High Performers (well-rounded, high performing interventions)}
        
        The second largest cluster included 12 interventions. These were mostly platform-implemented interventions that scored relatively well across all five metrics. It included many of the less intrusive or higher agency affording interventions like demonetization of user accounts (rather than suspensions or shadowbanning), limits on forwarding or resharing content, and several labeling and user-based measures like fact-checking labels and improved user reporting.
        
        \item \textit{Effective but Contested (effective and low-cost, but less popular interventions)}
        
        The largest group of interventions (15) consisted of platform-implemented controls that were rated as being low cost to implement and effective to deploy. However, they have relatively low user acceptance. Generally, this category includes many of the more restrictive or intrusive account moderation, content moderation, and content distribution interventions, such as account suspensions, downranking, and banning political ads.
        
        \item \textit{Effective but Costly (effective and acceptable, but resource-intensive interventions)}
    
        This category included 9 interventions and mostly consisted of almost all of the external institutional measures (such as investing in journalism, pursuing privacy legislation or anti-trust action) and the media literacy/educational measures. These interventions had by far the highest costs and effort associated with implementing and maintaining these programs, and ranked on the lower end of the spectrum for political feasibility relative to other clusters.
        
        \item \textit{Easy Starters (politically feasible and popular, but low-impact interventions) }
    
        The smallest cluster included only 4 interventions, and it included the measures with by far the highest overall user acceptance scores and highest likelihood of implementation (such as crowdsourcing context labels, regulating deepfakes, and regulating AI in ads). However, in general, they were viewed as less effective than most other measures. 
\end{enumerate}

\begin{table*}[t]
\centering
\footnotesize
\setlength{\tabcolsep}{3pt}
\begin{tabular}{lllcccccc|c}
\toprule
 & \textbf{Category} & \textbf{Intervention} & \textbf{N} & \textbf{Feasibility} & \textbf{Effectiveness} & \textbf{Acceptance} & \textbf{Cheapness} & \textbf{Ease} & \textbf{Overall} \\
\midrule
\multirow{12}{*}{\rotatebox[origin=c]{90}{\parbox{2.8cm}{\centering{\textit{Balanced high performers}}}}} & Account Moderation & Demonetization & 12 & 3.58 (1.31) & 4.00 (0.60) & 3.75 (1.06) & 2.92 (1.51) & 2.92 (1.24) & 3.43 (0.52) \\
 & Content Moderation & Algorithmic Changes & 11 & 3.64 (1.03) & 4.09 (0.54) & 3.55 (1.04) & 3.00 (1.26) & 2.55 (0.93) & 3.36 (0.54) \\
 & Content Distribution & Accuracy Prompts & 8 & 3.38 (1.51) & 3.62 (0.92) & 3.50 (1.07) & 2.88 (1.13) & 3.38 (1.06) & 3.35 (0.75) \\
 & Content Distribution & Limit Forwarding & 13 & 4.08 (0.86) & 3.69 (0.63) & 3.54 (1.05) & 4.31 (0.85) & 3.92 (1.04) & 3.91 (0.49) \\
 & Content Distribution & Limit Resharing & 9 & 4.00 (1.22) & 3.78 (0.44) & 3.00 (1.22) & 4.56 (0.53) & 4.22 (0.67) & 3.91 (0.45) \\
 & Content Labeling & Fact-check Labels & 15 & 3.53 (1.06) & 3.87 (0.74) & 3.80 (0.94) & 2.79 (1.12) & 2.21 (1.12) & 3.26 (0.59) \\
 & Content Labeling & Government Labels & 5 & 4.00 (0.00) & 3.80 (1.30) & 4.00 (0.00) & 4.40 (0.55) & 4.00 (0.00) & 4.04 (0.26) \\
  & Content Labeling & AI Disclosure & 14 & 3.71 (1.59) & 3.71 (0.99) & 4.07 (0.83) & 3.31 (1.25) & 3.31 (1.18) & 3.64 (0.75) \\
 & User-based Measures & Reporting & 10 & 4.40 (0.52) & 3.80 (0.79) & 4.00 (0.94) & 3.00 (1.49) & 2.40 (1.07) & 3.52 (0.45) \\
 & User-based Measures & Social Norms & 13 & 3.62 (0.77) & 3.62 (0.96) & 3.83 (0.94) & 3.33 (1.23) & 2.67 (1.37) & 3.40 (0.60) \\
 & User-based Measures & Alter Platform Metrics & 12 & 4.08 (0.79) & 3.92 (1.00) & 3.75 (1.36) & 3.00 (1.41) & 2.50 (1.09) & 3.45 (0.62) \\
 & Institutional Measures & Targeted Ads & 13 & 3.38 (1.19) & 3.38 (0.65) & 3.85 (0.80) & 3.08 (1.38) & 3.33 (1.30) & 3.44 (0.64) \\
\midrule
 & & \textit{Cluster 1 mean} & & \textbf{3.78} & \textbf{3.77} & \textbf{3.72} & \textbf{3.38} & \textbf{3.12} & \textbf{3.56}  \\
\midrule
\multirow{15}{*}{\rotatebox[origin=c]{90}{\parbox{3.5cm}{\centering{\textit{Effective and low-cost, but less popular interventions}}}}} & Account Moderation & Account Suspensions & 15 & 2.80 (1.52) & 3.53 (0.99) & 2.86 (1.29) & 3.47 (1.13) & 2.40 (0.83) & 3.02 (0.64) \\
 & Account Moderation & Deplatforming & 12 & 2.58 (1.31) & 3.83 (0.83) & 2.92 (1.00) & 3.25 (1.36) & 2.33 (1.07) & 2.98 (0.38) \\
 & Account Moderation & Shadow Banning & 9 & 3.44 (1.24) & 3.33 (1.32) & 2.89 (1.05) & 3.50 (1.69) & 3.38 (1.51) & 3.27 (0.90) \\
 & Content Moderation & Content Removal & 9 & 3.11 (1.17) & 3.89 (0.93) & 3.33 (0.87) & 3.78 (0.97) & 2.44 (1.24) & 3.31 (0.66) \\
 & Content Moderation & Downranking & 13 & 3.15 (1.46) & 3.92 (0.76) & 3.23 (1.17) & 3.15 (1.28) & 3.08 (1.04) & 3.31 (0.69) \\
 & Content Moderation & Virality Circuit Breakers & 12 & 3.58 (1.08) & 3.58 (1.38) & 3.08 (0.90) & 3.00 (1.28) & 2.92 (1.00) & 3.23 (0.65) \\
 & Content Distribution & Redirection & 10 & 2.60 (1.35) & 3.10 (0.99) & 3.10 (1.10) & 3.90 (0.88) & 3.10 (1.10) & 3.16 (0.45) \\
 & Content Distribution & Friction & 12 & 3.75 (0.97) & 3.50 (1.00) & 3.33 (1.07) & 3.83 (1.47) & 3.33 (1.15) & 3.55 (0.67) \\
 & Content Distribution & Platform Alterations & 10 & 3.60 (1.07) & 3.80 (1.32) & 3.00 (0.94) & 3.10 (1.37) & 2.50 (0.97) & 3.20 (0.65) \\
 & Content Distribution & Ban Political Ads & 15 & 2.20 (1.32) & 3.00 (0.93) & 3.21 (1.37) & 3.27 (1.49) & 3.60 (1.30) & 3.05 (0.72) \\
  & Generative AI & Gen AI Chatbots & 18 & 3.39 (0.92) & 3.24 (0.75) & 2.71 (0.92) & 2.65 (1.17) & 2.82 (1.01) & 2.96 (0.59) \\
 & Content Labeling & Warning Labels & 12 & 3.42 (1.08) & 3.75 (0.97) & 3.17 (1.03) & 3.17 (1.40) & 2.42 (1.08) & 3.18 (0.67) \\
 & Content Labeling & Source Labeling & 9 & 3.11 (1.17) & 3.67 (0.87) & 3.67 (0.87) & 3.56 (1.24) & 2.33 (1.00) & 3.27 (0.66) \\
 & Generative AI & Gen AI Content & 12 & 3.33 (0.65) & 3.00 (0.85) & 3.09 (0.83) & 2.64 (0.81) & 2.73 (0.79) & 2.96 (0.50) \\
 & Institutional Measures & Taxes / Fines & 13 & 2.15 (0.99) & 3.08 (1.26) & 3.92 (1.32) & 2.83 (1.64) & 2.25 (1.48) & 2.88 (0.77) \\
\midrule
 & & \textit{Cluster 2 mean} & & \textbf{3.08} & \textbf{3.48} & \textbf{3.17} & \textbf{3.27} & \textbf{2.78} & \textbf{3.16}\\
\midrule
\multirow{9}{*}{\rotatebox[origin=c]{90}{\parbox{2.5cm}{\centering{\textit{Effective and acceptable, but resource-intensive}}}}} & Content Distribution & Fact-Check Ads & 9 & 3.33 (1.00) & 3.44 (0.73) & 3.75 (0.71) & 2.00 (0.93) & 2.00 (0.93) & 2.95 (0.62) \\
 & Media Literacy & Digital Media Literacy & 9 & 3.78 (0.83) & 3.78 (1.20) & 4.11 (0.93) & 2.00 (1.12) & 2.00 (1.12) & 3.13 (0.51) \\
 & Media Literacy & Inoculation & 10 & 3.60 (0.84) & 3.40 (0.84) & 3.90 (0.88) & 2.50 (0.97) & 2.10 (0.99) & 3.10 (0.56) \\
 & Institutional Measures & Media Support & 10 & 3.30 (1.34) & 3.40 (1.07) & 3.78 (0.67) & 1.80 (0.92) & 2.00 (0.94) & 2.83 (0.66) \\
 & Institutional Measures & Journalism Support & 13 & 3.46 (1.20) & 4.00 (0.71) & 3.77 (1.01) & 1.38 (0.51) & 1.85 (0.80) & 2.89 (0.42) \\
 & Institutional Measures & Data Sharing & 11 & 3.00 (1.41) & 4.09 (0.83) & 4.18 (0.75) & 2.73 (1.01) & 2.00 (1.41) & 3.20 (0.86) \\
 & Institutional Measures & Government Regulation & 11 & 2.91 (1.14) & 3.64 (1.12) & 3.55 (0.82) & 1.82 (0.60) & 1.55 (1.04) & 2.69 (0.40) \\
 & Institutional Measures & Privacy Legislation & 12 & 2.75 (1.48) & 3.25 (1.36) & 3.80 (0.92) & 1.36 (0.50) & 1.09 (0.30) & 2.51 (0.86) \\
 & Institutional Measures & Anti-Trust Action & 10 & 2.20 (1.40) & 3.60 (1.07) & 4.20 (0.63) & 1.40 (0.70) & 1.70 (1.06) & 2.62 (0.55) \\
\midrule
 & & \textit{Cluster 3 mean} & & \textbf{3.15} & \textbf{3.62} & \textbf{3.89} & \textbf{1.89} & \textbf{1.81} & \textbf{2.88} \\
\midrule
\multirow{4}{*}{\rotatebox[origin=c]{90}{\parbox{1cm}{\centering\textit{Easy starters}}}} & Content Moderation & User Control & 14 & 4.21 (1.12) & 3.00 (1.18) & 4.36 (0.84) & 3.38 (1.39) & 2.31 (0.95) & 3.47 (0.70) \\
 & Content Distribution & AI in Ads & 18 & 3.61 (1.33) & 2.94 (1.00) & 3.89 (0.96) & 2.83 (1.34) & 3.00 (1.24) & 3.26 (0.68) \\
  & Content Moderation & Ban Deepfakes & 11 & 4.00 (0.89) & 2.55 (1.29) & 4.10 (0.99) & 3.00 (1.25) & 2.40 (1.17) & 3.26 (0.68) \\
 & Content Labeling & Crowdsourcing & 14 & 4.29 (0.61) & 3.21 (0.80) & 4.08 (0.64) & 4.14 (1.17) & 3.07 (1.27) & 3.76 (0.60) \\
\midrule
 & & \textit{Cluster 4 mean} & & \textbf{4.03} & \textbf{2.93} & \textbf{4.11} & \textbf{3.34} & \textbf{2.69} & \textbf{3.44}  \\
\bottomrule
\end{tabular}
\caption{Interventions sorted by cluster and category with mean scores and standard deviations (in parentheses). Cluster mean rows show the average across interventions in that cluster.}
\label{tab:ward-clusters}
\end{table*}

\section{Discussion}

In this study, 39 experts evaluated misinformation interventions based on five evaluative criteria: political feasibility, effectiveness, acceptance, cost, and effort. An intervention is never perfect; it is a trade-off between these evaluative criteria. The trade-offs between evaluative criteria for interventions are systematic and not random. Across both category-level and individual-level interventions, the data shows a recurring pattern, where interventions that are restrictive or opaque face acceptance and feasibility barriers regardless of their effectiveness. Informational and user-empowering interventions face the opposite problem of high acceptance but uncertain effectiveness. This finding should inform how platforms build their intervention portfolios, that they should diversify across multiple intervention types rather than maximizing on any single metric.

The main findings suggest that, like in many other public policy domains, people prefer informational interventions that afford them agency over more intrusive or restrictive ones \cite{diepeveenPublicAcceptabilityGovernment2013,hagmannTaxesLabelsNudges2018}. On average, content labeling, user-based measures, and content distribution interventions scored the highest among expert raters across the five metrics. Among these top three categories, content labeling interventions primarily target content after it has already spread, focusing on the belief phase of the misinformation pipeline. Content distribution interventions generally target the spread of misinformation, either addressing the direct sharing component (such as friction and accuracy prompts) or the amplification of the content (by limiting resharing or forwarding). The top-ranked user-based measures included improving user reporting, which also usually targets the spread phase. When determining which interventions to implement, these top categories unfortunately generally target only the \textbf{spread} and \textbf{belief} in misinformation rather than the \textbf{creation} of networks or content. An ideal enforcement strategy would be to determine and implement the best interventions across the misinformation pipeline. 

However, two moderation strategies ranked among the top 10 highest-ranked interventions: altering platform metrics to reward accuracy rather than engagement and using demonetization as an account moderation strategy. These interventions target the incentives associated with creating misinformation content and can also affect the potential spread and amplification. Experts in the elicitation survey ranked them among the most effective interventions (5th and 3rd, respectively), although they are perceived to have middling user acceptance and require a decent amount of effort or cost to implement. The literature supports these results, suggesting that reducing incentives to create and share misinformation (or conversely, increasing either monetary or social incentives to share accurate information) can considerably improve the quality of content shared and improve user discernment 
\cite{kapoorDoesIncentivizationPromote2023,globigChangingIncentiveStructure2023}.

Further, comparing our results with a similar survey of public opinion~\citep{altaySurveyExpertViews2023}, we note that expert judgment does not necessarily track public perception. For example, members of the public rate friction-based interventions, like delaying users from posting content they have not viewed, as very intrusive and among the least effective~\citep{kingPublicSupportMisinformation2026}, but experts rate them as moderately effective. Notably, this expert-public mismatch is not uniform. For content moderation, experts and the public converge on the trade-off structure itself, independently judging that both account and content moderation are effective yet unpopular. This observation argues for participatory design for sociotechnical interventions, where expert elicitation are not treated as a direct stand in for public input, but user opinions are also folded into the decision process. This would surface disagreements before deployment and create a more inclusive online space.

Finally, these sets of interventions go beyond the scope of misinformation. They target mechanisms of account and content visibility and distribution that recur across socio-technical systems more broadly. For example, account and content moderation are primary levers that platforms use to constrain the amplification of undesired narratives by automated and coordinated inauthentic accounts, regardless of whether the content is misinformation or not~\citep{ng2026fate}. 

Consider the case of coordinated inauthentic behavior, where networks of automated and human-operated accounts amplify desired narratives~\citep{ng2022online}. The same five criteria apply, and they reproduce the same tension we find in misinformation. Account-level takedowns are effective, but opaque and contested~\citep{ng2026fate}, placing them in the low-acceptance, high-effectiveness cluster; transparency measures such as behavioral labels are acceptable but effects may vary, echoing the easy-starter cluster. The lifecycle gap recurs here too. Takedowns act only after a network has already amplified content, leaving account creation, the stage at which coordination is the easiest to disrupt, largely unaddressed, because it is difficult to predict if the account will later coordinate. Generally, the lifecycle lens and the trade-off structure carry over, though the interventions may need re-fitting to the domain.


\subsection{Design Implications} 
Based on the main findings and the four calculated intervention clusters, we derive design implications and then propose a decision framework for practitioners.


\subsubsection{Intervention Cluster Implications}

The four calculated intervention clusters point to specific design implications. See Section \ref{sec:top_interventions} and Table \ref{tab:ward-clusters} for details on each of the four clusters.

\paragraph{\textbf{(1) Balanced High Performers - Design for user agency where the effectiveness case allows for it.}} 

Across the category-level comparison in Figure \ref{fig:heatmap} and the intervention-level analysis (Table \ref{tab:ward-clusters}), user agency is a recurring feature of the most acceptable platform interventions in the cluster of well-balanced high performers. The interventions that users find most acceptable preserve their ability to interpret, contest, or act on information, rather than removing the choice for them. Among the top 10 overall interventions (Table \ref{tab:intervention_rankings}),  labeling government sources, crowdsourcing labels, AI disclosure, encouraging user reporting, and allowing users to control their news feeds directly all preserve a high level of information or agency for the user, rather than restrict it. 

\paragraph{\textbf{(2) Effective but Contested - Acceptance should be a necessary design target, not cost.} }

Some of the most effective interventions, including many that are low-cost and have a low administrative burden, are unfortunately not always acceptable to users. As shown in Figure \ref{fig:general_eff_acc}, large majorities of our expert participants rated the account and content moderation categories as effective (74\% and 77\%, respectively), but they also show the largest drop in percentage that believe these interventions would be acceptable to the general public (only 38\% and 54\%, respectively). This gap is also acknowledged in public policy literature, where acceptance is driven by the perceived fairness, transparency and trust in the implementation, or the implementer, rather than the intervention's actual effectiveness~\citep{grelleWhenWhyPeople2024}.
Therefore, implementers should focus on increasing user acceptance by improving trust in the systems and also improving the public messaging.
Prior HCI work also shows that users tend to distrust what they cannot see or appeal~\citep{myers2018censored,ng2026fate}, so labeling systems should reduce opacity and include recourse and source transparency.

\paragraph{\textbf{(3) Effective but Costly - Lower the cost of interventions that are already rated as highly effective and acceptable.} }

The media literacy and external institutional measures in cluster 3 are rated as both highly effective and among the most acceptable, yet their overall scores on average are the lowest (see Table \ref{tab:ward-clusters}, Figure \ref{fig:heatmap}) because they are often too resource-intensive to be practiced. In this case, the trade-off is not values-based (as in moderation), but the burden of implementation. This points to a different design lever: reducing delivery cost rather than reducing scope. Meta-analytic work on media literacy interventions show that effectiveness is a measure that is sensitive to how and where an intervention is delivered~\citep{huangMediaLiteracyInterventions2024}, suggesting that institutions without dedicated funding should look to amortize these interventions into existing infrastructure. For example, literacy prompts could be embedded into product onboarding flows rather than a freestanding curricular. 

\paragraph{\textbf{(4) Easy Starters - Emerging interventions that are straightforward and low-cost are a good first step. }}

Cluster 4 includes many newer interventions, such as regulating AI in social media ads, controlling the use of deepfakes, crowdsourcing fact-checking labels, and giving users control over their news feeds. These interventions have less evidence of effectiveness than the more established interventions in clusters 1-3, and their implementation would likely require creating and deploying new features. As Table \ref{tab:ward-clusters} shows, these interventions score poorly on effectiveness and ease of implementation but rank highest among all clusters in political feasibility and user acceptance. This may partly be because these interventions are either high-agency-preserving or focus on regulating or controlling AI, a technology which is currently relatively unpopular across demographic groups in the US~\citep{mcclainHowUSPublic2025}. Since they are also low-cost and relatively simple to deploy, requiring only that platforms choose to do so, they present a low-risk option that platforms could adopt. Implementers can quickly adopt them, monitor user responses, and iterate to improve these interventions. These changes could enhance their future effectiveness and acceptability. We therefore recommend that practitioners view cluster 4 not as a final set of solutions but as a starting point: low-cost interventions that can be deployed quickly, are transparent, and empower users. This can help establish the legitimacy and trust needed between users and platforms to later implement more restrictive yet more effective interventions.

\subsubsection{General Implications}
\label{sec:gen_implications}
Beyond the cluster-level patterns, a structural gap emerges when interventions are mapped onto the information lifecycle (Figure \ref{fig:information_lifecycle}). The most viable interventions, i.e. those in Cluster 1, are concentrated at the spread and belief phases, leaving the creation phase largely unaddressed. 
Of the top 10 interventions overall (Table \ref{tab:intervention_rankings}), only two (altering platform metrics, demonetization), directly target the incentives that drive content creation. However, both have middling acceptance scores (both average a 3.75 on a 5-pt Likert scale, and rank 19th/ 20th out of 40 interventions on acceptance respectively). This aligns with research showing that misinformation persists because of platform incentive structures that favor engagement over accuracy at the point of content creation~\citep{globigChangingIncentiveStructure2023}. This is not merely an artifact of what has been studied, instead, it reflects a structural feature of the field: the most mature, acceptable and deployable tools act downstream of the incentive structures that generate the content in the first place, and are unable to address the root cause of content creation~\citep{globigChangingIncentiveStructure2023}.

This pipeline coverage gap has direct implications for platform architecture. Closing it will require designing creation-phase interventions that can occupy the same viable space as the spread and belief phase interventions in Cluster 1 -- transparency, agency-preserving and low-burden interventions. For the existing creation-phase interventions that come the closest (demonetization and altering platform metrics), they are already pointing in this direction: these interventions work by restructuring incentives rather than restricting behavior, which likely explains why they are better received than more overtly restrictive creation-phase measures like account suspension and deplatforming. 

It is critical that all stages of the information lifecycle are addressed with interventions. Future intervention design should take this as a signal: creation-phase interventions framed as restructuring incentives, rather than restricting participation, may be more likely to achieve the acceptance needed for sustained deployment.
More broadly, these findings argue against single-intervention strategies, and that combining interventions across different mechanisms reduces the spread of viral misinformation~\citep{bak-colemanCombiningInterventionsReduce2022}. Because the most viable interventions cluster at the specific pipeline stages, and trade-off against each other on the five criteria, an effective implementation strategy requires a portfolio that deliberately spans the lifecycle. It should pair high-acceptance spread and belief-phase interventions with creation-phase measures whose acceptance can be incrementally built. In the next section, we propose a decision framework that is intended to support this kind of portfolio reasoning.




\subsubsection{A Decision Framework} 
The four empirically derived clusters translate directly into a decision framework for practitioners. Organizations deploying socio-technical interventions typically operate under a specific set of constraints (i.e., budget, political exposure, timeline, or public trust), and the clusters align cleanly with the four constraint profiles. Table~\ref{tab:decision-framework} maps each cluster to the conditions under which it is the appropriate starting point, with representative interventions and design considerations. This proposed decision framework should be used as a navigation tool. It begins with where the practitioner is constrained, which identifies the starting cluster of interventions, then, what design investments are needed to unlock the next cluster.

A key implication of this framework is that constraint profiles change over time, and the clusters are designed to be used sequentially as well as independently. Cluster 4 interventions (that are low-cost, high user acceptance and politically feasible), are most appropriate as a first step to build the legitimacy of the interventions, establishing the trust between users and platforms. This is a precondition for later deploying more effective but more contested interventions that are in Clusters 1 and 2. Cluster 3 interventions are both highly effective and well-received but are resource-intensive, and therefore should be pursued when coalition partners or dedicated funding can amortize the delivery cost. For example, embedding media literacy prompts into product onboarding flows rather than standalone curricula, is one way to reduce the structural barrier without reducing the intervention scope.

Critically, no single cluster is sufficient on its own. The pipeline coverage gap identified in Section~\ref{sec:gen_implications} means that an intervention portfolio needs to balance across multiple clusters. For example, sole reliance on Cluster 1 interventions will systematically under-address account and content creation. Therefore, practitioners should pair any Cluster 1 deployment with at least one creation-phase measure, most likely from Cluster 2. They should also invest in the transparency and recourse mechanisms that can shift the measure's acceptance over time.

\begin{table*}[t]
  \caption{A practitioner decision framework for selecting misinformation
    interventions, mapped to the four empirically derived clusters in
    Table~\ref{tab:ward-clusters}. $^\dagger$ marks creation-phase interventions.}
  \label{tab:decision-framework}
  \centering
  \begin{tabular}{@{}
      >{\raggedright\arraybackslash}p{0.16\textwidth}
      >{\raggedright\arraybackslash}p{0.21\textwidth}
      >{\raggedright\arraybackslash}p{0.24\textwidth}
      >{\raggedright\arraybackslash}p{0.32\textwidth}
    @{}}
    \toprule
    \textbf{Cluster} & \textbf{When to choose it} & \textbf{Example interventions} & \textbf{Design considerations} \\
    \midrule
    \textbf{1}\enspace Well-balanced, high-performing &
    Needs broadly viable wins without requiring unusual budget or political cover &
    Demonetization$^\dagger$; limits on resharing/forwarding; fact-check and government labels; reporting; AI disclosure labels &
    Preserves users' ability to interpret, contest, or act on information rather than removing the choice for them.
    Pair with at least one creation-phase measure from cluster 2 and invest in increasing that measure's acceptance \\
    \midrule
    \textbf{2}\enspace Effective and affordable, but low acceptance &
    Has an enforcement mandate or legal cover and can absorb public pushback &
    Account suspension$^\dagger$; deplatforming$^\dagger$; shadow banning$^\dagger$; downranking; content removal &
    Acceptance cost is addressable, not fixed. Invest in transparent criteria, reduced opacity, recourse options, and error minimization \\
    \midrule
    \textbf{3}\enspace Effective and acceptable, but resource-intensive &
    Has a long time horizon, coalition partners, or dedicated funding &
    Media literacy programs; journalism/media support; data sharing; regulation &
    Share cost across a coalition (platform + civil society + government); look for delivery channels that amortize costs \\
    \midrule
    \textbf{4}\enspace Emerging, straightforward, and low-cost &
    Needs to build public trust or political capital before bigger moves &
    Crowdsourcing; user control; deepfake/AI-in-ads regulation &
    Useful as a legitimacy-building first step or complement, designed with user agency in mind. Treat as an initial testbed to iterate and improve effectiveness, not as the primary countermeasure \\
    \bottomrule
  \end{tabular}
\end{table*}

\subsection{Limitations}
While this study is among the first expert surveys on misinformation interventions and the only one so far to analyze multiple evaluative metrics at once, several limitations are noted. First, our sample was limited in both size (n = 39) and geographical reach. The researchers surveyed were based predominantly at a \textbf{Northern American academic institution, with researchers from [[Host university name withheld]]} in particular representing a high fraction of respondents.
Second, while the intervention list had forty interventions, respondents only each rated twelve of the interventions to limit the survey length and improve response rates. This limited the sample size for each specific intervention, although all respondents rated the overall effectiveness and acceptance of the general countermeasures categories. Future work should consider a larger sample size or a more diverse set of respondents. We additionally weighted all five evaluative criteria equally, although some are plausibly correlated. An extension of this work could analyze the relationship between the metrics and how that may affect overall ratings. Further, there may be metrics not included in this work that may be relevant to practitioners or policymakers that could be valuable to investigate in future research. Finally, this survey captures expert judgment only, and not public perception of interventions. While our discussion compares our expert ratings against companion user findings from literature, that comparison spans separate data collection efforts. Future work should expand the survey to the general public for a paired measurement.

\section{Conclusion}
This work set out to give designers and policymakers a principled basis for comparing sociotechnical interventions, rather than evaluating them one at a time on effectiveness alone. We defined eight categories of interventions and forty operationalized interventions that can be deployed along the information lifecycle. Instantiating this through the case of misinformation, we surveyed $N=39$ researchers on the forty interventions across five evaluative criteria. The five criteria are: political feasibility, effectiveness, acceptance, cost, and effort. Our results find that interventions that experts judge to be the most effective are often not those judged to be most acceptable or feasible. User-based measures, content labeling and content distribution measures scored the highest in multiple evaluative metrics. Additionally, some provably effective interventions, such as content and account moderation techniques, lack user support and acceptance due to their perceived unfairness or intrusiveness. 

An intervention that experts rate highly effective can still fail in deployment if it is unacceptable, too costly or politically infeasible. 
The framework we presented is a first step towards a comparative, multi-criteria evaluation of sociotechnical interventions. The framework is intended to support principled design decisions by policymakers and platform governance teams through identifying trade-offs.
This framework also extends naturally beyond misinformation, including to interventions targeting the automated coordinated accounts that platforms moderate independently of content. The five criteria, the lifecycle lens and the cluster logic separates viable-but-limited interventions from effective-but-contested ones, providing a repeatable method for designing and evaluating such socio-technical interventions in different scenarios. 

\begin{acks}
This work was supported by the Knight Foundation, the Office of Naval Research’s MURI: Persuasion, Identity, \& Morality in Social-Cyber Environments grant N00014-21-12749, Carnegie Mellon University’s Graduate small Project Help (GuSH), the Center for Computational Analysis of Social and Organizational Systems (CASOS), and the Center for Informed Democracy and Social-cybersecurity (IDeaS). The views and conclusions contained in this document are those of the authors alone. The funders have no role in study design, data collection and analysis, decision to publish, or preparation of the manuscript. 
\end{acks}

\bibliographystyle{ACM-Reference-Format}
\bibliography{references}

@article{blairInterventionsCounterMisinformation2024,
	title = {Interventions to counter misinformation: {Lessons} from the {Global} {North} and applications to the {Global} {South}},
	volume = {55},
	issn = {2352-250X},
	shorttitle = {Interventions to counter misinformation},
	url = {https://www.sciencedirect.com/science/article/abs/pii/S2352250X2300177X},
	doi = {10.1016/j.copsyc.2023.101732},
	urldate = {2024-03-19},
	journal = {Current Opinion in Psychology},
	author = {Blair, Robert A. and Gottlieb, Jessica and Nyhan, Brendan and Paler, Laura and Argote, Pablo and Stainfield, Charlene J.},
	month = feb,
	year = {2024},
	pages = {101732}
}

@article{courchesneReviewSocialScience2021,
	title = {Review of social science research on the impact of countermeasures against influence operations},
	volume = {2},
	doi = {10.37016/mr-2020-79},
	urldate = {2021-09-16},
	journal = {Harvard Kennedy School Misinformation Review},
	author = {Courchesne, Laura and Ilhardt, Julia and Shapiro, Jacob N.},
	month = sep,
	year = {2021}
}

@article{kingPublicSupportMisinformation2026,
	title = {Public {Support} for {Misinformation} {Interventions} {Depends} {On} {Perceived} {Fairness}, {Effectiveness}, and {Intrusiveness}},
	volume = {3},
	issn = {2770-3142},
	url = {https://tsjournal.org/index.php/jots/article/view/267},
	doi = {10.54501/jots.v3i2.267},
	number = {2},
	urldate = {2026-06-22},
	journal = {Journal of Online Trust and Safety},
	author = {King, Catherine and Phillips, Samantha C. and Carley, Kathleen M.},
	month = mar,
	year = {2026}
}

@article{kozyrevaToolboxIndividuallevelInterventions2024,
	title = {Toolbox of individual-level interventions against online misinformation},
	copyright = {2024 Springer Nature Limited},
	issn = {2397-3374},
	url = {https://www.nature.com/articles/s41562-024-01881-0},
	doi = {10.1038/s41562-024-01881-0},
	urldate = {2024-05-14},
	journal = {Nature Human Behaviour},
	author = {Kozyreva, Anastasia and Lorenz-Spreen, Philipp and Herzog, Stefan M. and Ecker, Ullrich K. H. and Lewandowsky, Stephan and Hertwig, Ralph and Ali, Ayesha and Bak-Coleman, Joe and Barzilai, Sarit and Basol, Melisa and Berinsky, Adam J. and Betsch, Cornelia and Cook, John and Fazio, Lisa K. and Geers, Michael and Guess, Andrew M. and Huang, Haifeng and Larreguy, Horacio and Maertens, Rakoen and Panizza, Folco and Pennycook, Gordon and Rand, David G. and Rathje, Steve and Reifler, Jason and Schmid, Philipp and Smith, Mark D. and Swire-Thompson, Briony and Szewach, Paula and van der Linden, Sander and Wineburg, Sam},
	month = may,
	year = {2024},
	pages = {1044–-1052},
    volume = {8},
}

@article{altaySurveyExpertViews2023,
	title = {A survey of expert views on misinformation: {Definitions}, determinants, solutions, and future of the field},
	shorttitle = {A survey of expert views on misinformation},
	url = {https://misinforeview.hks.harvard.edu/article/a-survey-of-expert-views-on-misinformation-definitions-determinants-solutions-and-future-of-the-field/},
	doi = {10.37016/mr-2020-119},
    volume = {4},
    issue = {4},
	language = {en-US},
	urldate = {2023-08-01},
	journal = {Harvard Kennedy School Misinformation Review},
	author = {Altay, Sacha and Berriche, Manon and Heuer, Hendrik and Farkas, Johan and Rathje, Steven},
	month = jul,
	year = {2023}
}

@book{dudleyRegulationPrimer2012,
	address = {Arlington, Va},
	edition = {2nd ed},
	title = {Regulation: {A} {Primer}},
	isbn = {978-0-9836077-3-1 978-0-9836077-4-8},
	shorttitle = {Regulation},
	publisher = {Mercatus Center at George Mason University},
	editor = {Dudley, Susan E. and Brito, Jerry},
	year = {2012}
}

@article{grelleWhenWhyPeople2024,
	title = {When and {Why} {Do} {People} {Accept} {Public}-{Policy} {Interventions}? {An} {Integrative} {Public}-{Policy}-{Acceptance} {Framework}},
	volume = {19},
	issn = {1745-6916},
	shorttitle = {When and {Why} {Do} {People} {Accept} {Public}-{Policy} {Interventions}?},
	url = {https://doi.org/10.1177/17456916231180580},
	doi = {10.1177/17456916231180580},
	language = {en},
	number = {1},
	urldate = {2025-02-05},
	journal = {Perspectives on Psychological Science},
	author = {Grelle, Sonja and Hofmann, Wilhelm},
	month = jan,
	year = {2024},
	pages = {258--279}
}

@book{kraftPublicPolicyPolitics2017,
	edition = {6th},
	title = {Public {Policy}: {Politics}, {Analysis}, and {Alternatives}},
	isbn = {978-1-5063-5814-7},
	shorttitle = {Public {Policy}},
	language = {en},
	publisher = {CQ Press},
	author = {Kraft, Michael E. and Furlong, Scott R.},
	month = apr,
	year = {2017}
}

@article{rochefortRegulatingSocialMedia2020,
	title = {Regulating {Social} {Media} {Platforms}: {A} {Comparative} {Policy} {Analysis}},
	volume = {25},
	issn = {1081-1680, 1532-6926},
	shorttitle = {Regulating {Social} {Media} {Platforms}},
	url = {https://www.tandfonline.com/doi/full/10.1080/10811680.2020.1735194},
	doi = {10.1080/10811680.2020.1735194},
	number = {2},
	urldate = {2022-10-17},
	journal = {Communication Law and Policy},
	author = {Rochefort, Alex},
	month = apr,
	year = {2020},
	pages = {225--260}
}

@article{diepeveenPublicAcceptabilityGovernment2013,
	title = {Public acceptability of government intervention to change health-related behaviours: a systematic review and narrative synthesis},
	volume = {13},
	issn = {1471-2458},
	shorttitle = {Public acceptability of government intervention to change health-related behaviours},
	url = {https://doi.org/10.1186/1471-2458-13-756},
	doi = {10.1186/1471-2458-13-756},
	number = {1},
	urldate = {2025-02-05},
	journal = {BMC Public Health},
	author = {Diepeveen, Stephanie and Ling, Tom and Suhrcke, Marc and Roland, Martin and Marteau, Theresa M.},
	month = aug,
	year = {2013},
	pages = {756}
}

@article{hagmannTaxesLabelsNudges2018,
	title = {Taxes, labels, or nudges? {Public} acceptance of various interventions designed to reduce sugar intake},
	volume = {79},
	issn = {0306-9192},
	shorttitle = {Taxes, labels, or nudges?},
	url = {https://www.sciencedirect.com/science/article/pii/S0306919217310096},
	doi = {10.1016/j.foodpol.2018.06.008},
	urldate = {2025-02-05},
	journal = {Food Policy},
	author = {Hagmann, Désirée and Siegrist, Michael and Hartmann, Christina},
	month = aug,
	year = {2018},
	pages = {156--165}
}

@article{reynoldsCommunicatingEffectivenessIneffectiveness2020,
	title = {Communicating the effectiveness and ineffectiveness of government policies and their impact on public support: a systematic review with meta-analysis},
	volume = {7},
	shorttitle = {Communicating the effectiveness and ineffectiveness of government policies and their impact on public support},
	url = {https://doi.org/10.1098/rsos.190522},
	doi = {10.1098/rsos.190522},
	number = {1},
	urldate = {2025-02-05},
	journal = {Royal Society Open Science},
	author = {Reynolds, J. P. and Stautz, K. and Pilling, M. and van der Linden, S. and Marteau, T. M.},
	month = jan,
	year = {2020},
	pages = {190522}
}

@techreport{greenEvidenceBasedMisinformationInterventions2022,
	title = {Evidence-{Based} {Misinformation} {Interventions}: {Challenges} and {Opportunities} for {Measurement} and {Collaboration}},
	shorttitle = {Evidence-{Based} {Misinformation} {Interventions}},
	url = {https://carnegieendowment.org/2023/01/09/evidence-based-misinformation-interventions-challenges-and-opportunities-for-measurement-and-collaboration-pub-88661},
	language = {en},
	urldate = {2023-07-07},
	institution = {Carnegie Endowment for International Peace},
	author = {Green, Yasmin and Gully, Andrew and Roth, Yoel and Roy, Abhishek and Tucker, Joshua A. and Wanless, Alicia},
	month = dec,
	year = {2022}
}

@article{jiangTradeoffcenteredFrameworkContent2023,
	title = {A {Trade}-off-centered {Framework} of {Content} {Moderation}},
	volume = {30},
	issn = {1073-0516},
	url = {https://dl.acm.org/doi/10.1145/3534929},
	doi = {10.1145/3534929},
	articleno = {3}, 
    numpages = {34},
	urldate = {2023-05-03},
	journal = {ACM Transactions on Computer-Human Interaction},
	author = {Jiang, Jialun Aaron and Nie, Peipei and Brubaker, Jed R. and Fiesler, Casey},
	month = mar,
	year = {2023}
}

@article{huangMediaLiteracyInterventions2024,
	title = {Media {Literacy} {Interventions} {Improve} {Resilience} to {Misinformation}: {A} {Meta}-{Analytic} {Investigation} of {Overall} {Effect} and {Moderating} {Factors}},
	issn = {0093-6502},
	shorttitle = {Media {Literacy} {Interventions} {Improve} {Resilience} to {Misinformation}},
	url = {https://doi.org/10.1177/00936502241288103},
	doi = {10.1177/00936502241288103},
	urldate = {2025-02-20},
	journal = {Communication Research},
	author = {Huang, Guanxiong and Jia, Wufan and Yu, Wenting},
	month = oct,
	year = {2024},
	pages = {00936502241288103}
}

@article{maertensLongtermEffectivenessInoculation2021,
	title = {Long-term effectiveness of inoculation against misinformation: {Three} longitudinal experiments},
	volume = {27},
	issn = {1939-2192},
	shorttitle = {Long-term effectiveness of inoculation against misinformation},
	doi = {10.1037/xap0000315},
	number = {1},
	journal = {Journal of Experimental Psychology: Applied},
	author = {Maertens, Rakoen and Roozenbeek, Jon and Basol, Melisa and van der Linden, Sander},
	year = {2021},
	pages = {1--16}
}

@article{porterGlobalEffectivenessFactchecking2021,
	title = {The global effectiveness of fact-checking: {Evidence} from simultaneous experiments in {Argentina}, {Nigeria}, {South} {Africa}, and the {United} {Kingdom}},
	volume = {118},
	issn = {0027-8424, 1091-6490},
	shorttitle = {The global effectiveness of fact-checking},
	url = {https://pnas.org/doi/full/10.1073/pnas.2104235118},
	doi = {10.1073/pnas.2104235118},
	number = {37},
	urldate = {2023-07-07},
	journal = {Proceedings of the National Academy of Sciences},
	author = {Porter, Ethan and Wood, Thomas J.},
	month = sep,
	year = {2021},
	pages = {e2104235118}
}

@article{zhouEffectiveReportingSystem2024,
	title = {Effective reporting system to encourage users’ reporting behavior in social media platforms: an empirical study based on structural empowerment theory},
	volume = {43},
	issn = {0144-929X},
	shorttitle = {Effective reporting system to encourage users’ reporting behavior in social media platforms},
	url = {https://doi.org/10.1080/0144929X.2023.2281491},
	doi = {10.1080/0144929X.2023.2281491},
	number = {14},
	urldate = {2025-01-17},
	journal = {Behaviour \& Information Technology},
	author = {Zhou, Hong and Lu, Yaobin and Zhao, Ling and Wang, Bin and Li, Ting},
	month = oct,
	year = {2024},
	pages = {3490--3509}
}

@techreport{DemocracyDesignSocial2024,
	title = {Democracy by {Design}: {Social} {Media}’s {Policy} {Scores}},
	shorttitle = {Democracy by {Design}},
	url = {https://accountabletech.org/research/democracy-by-design-social-medias-policy-scores/},
	language = {en-US},
	urldate = {2025-01-03},
	year = {2024},
    institution = {Accountable Tech},
    author       = {{Accountable Tech}}
}

@misc{kaplanMoreSpeechFewer2025,
	title = {More {Speech} and {Fewer} {Mistakes}},
	url = {https://about.fb.com/news/2025/01/meta-more-speech-fewer-mistakes/},
	language = {en-US},
	urldate = {2025-03-04},
	journal = {Meta},
	author = {Kaplan, Joel},
	month = jan,
	year = {2025}
}

@misc{vogelsSupportMoreRegulation2022,
	title = {Support for more regulation of tech companies has declined in {U}.{S}., especially among {Republicans}},
	url = {https://www.pewresearch.org/short-reads/2022/05/13/support-for-more-regulation-of-tech-companies-has-declined-in-u-s-especially-among-republicans/},
	language = {en-US},
	urldate = {2025-01-02},
	journal = {Pew Research Center},
	author = {Vogels, Emily A.},
	month = may,
	year = {2022}
}

@misc{randAmericansActuallyWant2025,
	title = {Americans actually do want expert content moderation},
	url = {https://thehill.com/opinion/technology/5109667-sorry-zuckerberg-americans-actually-do-want-expert-content-moderation/},
	urldate = {2025-01-29},
	journal = {The Hill},
	author = {Rand, David and Martel, Cameron},
	month = jan,
	year = {2025},
}

@misc{zhanHeresWhyMeta2025,
	title = {Here's why {Meta} ended fact-checking, according to experts},
	url = {https://abcnews.go.com/US/why-did-meta-remove-fact-checkers-experts-explain/story?id=117417445},
	urldate = {2025-03-08},
	journal = {ABC News},
	author = {Zhan, Max},
	month = jan,
	year = {2025}
}

@article{globigChangingIncentiveStructure2023,
	title = {Changing the incentive structure of social media platforms to halt the spread of misinformation},
	volume = {12},
	issn = {2050-084X},
	url = {https://doi.org/10.7554/eLife.85767},
	doi = {10.7554/eLife.85767},
	urldate = {2025-05-01},
	journal = {eLife},
	author = {Globig, Laura K and Holtz, Nora and Sharot, Tali},
	editor = {Gillan, Claire M and Frank, Michael J and Gillan, Claire M},
	month = jun,
	year = {2023},
	pages = {e85767}
}

@article{kapoorDoesIncentivizationPromote2023,
	title = {Does incentivization promote sharing “true” content online?},
	url = {https://misinforeview.hks.harvard.edu/article/does-incentivization-promote-sharing-true-content-online/},
	doi = {10.37016/mr-2020-120},
	language = {en-US},
	urldate = {2025-05-01},
	journal = {Harvard Kennedy School Misinformation Review},
	author = {Kapoor, Hansika and Rezaei, Sarah and Gurjar, Swanaya and Tagat, Anirudh and George, Denny and Budhwar, Yash and Puthillam, Arathy},
	month = aug,
	year = {2023},
    volume = {4},
    number = {4}
}

@inproceedings{schoenebeck2023online,
  title={Online harassment in majority contexts: Examining harms and remedies across countries},
  author={Schoenebeck, Sarita and Batool, Amna and Do, Giang and Darling, Sylvia and Grill, Gabriel and Wilkinson, Daricia and Khan, Mehtab and Toyama, Kentaro and Ashwell, Louise},
  booktitle={Proceedings of the 2023 CHI conference on human factors in computing systems},
  pages={1--16},
  year={2023}
}

@inproceedings{selbst2019fairness,
  title={Fairness and abstraction in sociotechnical systems},
  author={Selbst, Andrew D and Boyd, Danah and Friedler, Sorelle A and Venkatasubramanian, Suresh and Vertesi, Janet},
  booktitle={Proceedings of the conference on fairness, accountability, and transparency},
  pages={59--68},
  year={2019}
}

@book{friedman2019value,
  title={Value sensitive design: Shaping technology with moral imagination},
  author={Friedman, Batya and Hendry, David G},
  year={2019},
  publisher={Mit Press}
}

@book{gillespie2018custodians,
  title={Custodians of the Internet: Platforms, content moderation, and the hidden decisions that shape social media},
  author={Gillespie, Tarleton},
  year={2018},
  publisher={Yale University Press}
}

@article{varela2025countermeasures,
  title={Countermeasures against fake news: a Delphi study},
  author={Varela da Costa, Jo{\~a}o and Mira da Silva, Miguel},
  journal={Transforming Government: People, Process and Policy},
  volume={19},
  number={2},
  pages={392--413},
  year={2025},
  publisher={Emerald Publishing Limited}
}

@article{reuter2025combating,
  title={Combating information warfare: state and trends in user-centred countermeasures against fake news and misinformation},
  author={Reuter, Christian and Lee Hughes, Amanda and Buntain, Cody},
  journal={Behaviour \& Information Technology},
  volume={44},
  number={13},
  pages={3348--3361},
  year={2025},
  publisher={Taylor \& Francis}
}

@article{acquisti2017nudges,
  title={Nudges for privacy and security: Understanding and assisting users’ choices online},
  author={Acquisti, Alessandro and Adjerid, Idris and Balebako, Rebecca and Brandimarte, Laura and Cranor, Lorrie Faith and Komanduri, Saranga and Leon, Pedro Giovanni and Sadeh, Norman and Schaub, Florian and Sleeper, Manya and others},
  journal={ACM Computing Surveys (CSUR)},
  volume={50},
  number={3},
  pages={1--41},
  year={2017},
  publisher={ACM New York, NY, USA}
}

@article{mathur2019dark,
  title={Dark patterns at scale: Findings from a crawl of 11K shopping websites},
  author={Mathur, Arunesh and Acar, Gunes and Friedman, Michael J and Lucherini, Eli and Mayer, Jonathan and Chetty, Marshini and Narayanan, Arvind},
  journal={Proceedings of the ACM on human-computer interaction},
  volume={3},
  number={CSCW},
  pages={1--32},
  year={2019},
  publisher={ACM New York, NY, USA}
}

@book{o2006uncertain,
  title={Uncertain judgements: eliciting experts' probabilities},
  author={O'Hagan, Anthony and Buck, Caitlin E and Daneshkhah, Alireza and Eiser, J Richard and Garthwaite, Paul H and Jenkinson, David J and Oakley, Jeremy E and Rakow, Tim},
  year={2006},
  publisher={John Wiley \& Sons}
}

@article{wright2022automated,
  title={Automated platform governance through visibility and scale: On the transformational power of automoderator},
  author={Wright, Lucas},
  journal={Social Media+ Society},
  volume={8},
  number={1},
  pages={20563051221077020},
  year={2022},
  publisher={SAGE Publications Sage UK: London, England}
}

@article{ng2021does,
  title={How does fake news spread? Understanding pathways of disinformation spread through APIs},
  author={Ng, Lynnette HX and Taeihagh, Araz},
  journal={Policy \& Internet},
  volume={13},
  number={4},
  pages={560--585},
  year={2021},
  publisher={Wiley Online Library}
}

@article{king2025path,
  title={A path forward on online misinformation mitigation based on current user behavior},
  author={King, Catherine and Phillips, Samantha C and Carley, Kathleen M},
  journal={Scientific Reports},
  volume={15},
  number={1},
  pages={9475},
  year={2025},
  publisher={Nature Publishing Group UK London}
}

@article{ng2026fate,
  title={FATe of Bots: Ethical Considerations of Social Bot Detection},
  author={Ng, Lynnette Hui Xian and Pan, Ethan and Yoder, Michael and Carley, Kathleen},
  journal={ACM Journal on Responsible Computing},
  volume={3},
  number={2},
  pages={1--35},
  year={2026},
  publisher={ACM New York, NY}
}

@phdthesis{king2025thesis,
  author     = {King, Catherine},
  title      = {Effective and Practical Strategies for Combatting Misinformation},
  school     = {Carnegie Mellon University},
  year       = {2025},
  address    = {Pittsburgh, PA, USA},
  month      = {May},
  doi        = {10.1184/R1/29316179},
  type       = {{PhD Dissertation}} 
}

@inproceedings{king2025brims,
  title={Promoting Social Corrections: A Media Literacy Intervention for Misinformation on Social Media},
  author={King, Catherine and Carley, Kathleen M},
  booktitle={International Conference on Social Computing, Behavioral-Cultural Modeling and Prediction and Behavior Representation in Modeling and Simulation},
  editor={Thomson, Robert and Renshaw, Scott and Al-khateeb, Samer and Burger, Annetta and Park, Patrick and Pyke, Aryn A.},
  pages={223--232},
  year={2025},
  publisher={Springer Nature}
}

@inproceedings{aghajariReviewingInterventionsAddress2023,
	series = {{CSCW}},
	title = {Reviewing {Interventions} to {Address} {Misinformation}: {The} {Need} to {Expand} {Our} {Vision} {Beyond} an {Individualistic} {Focus}},
	volume = {7},
	shorttitle = {Reviewing {Interventions} to {Address} {Misinformation}},
	url = {https://doi.org/10.1145/3579520},
	doi = {10.1145/3579520},
	urldate = {2024-03-19},
	booktitle = {Proceedings of the {ACM} on {Human}-{Computer} {Interaction}},
	publisher = {Association for Computing Machinery},
	author = {Aghajari, Zhila and Baumer, Eric P. S. and DiFranzo, Dominic},
	month = apr,
	year = {2023},
	pages = {87:1--87:34}
}

@article{king2025icwsm,
	title = {Mapping the {Scientific} {Literature} on {Misinformation} {Interventions}: {A} {Bibliometric} {Review}},
	volume = {2025},
	shorttitle = {Mapping the {Scientific} {Literature} on {Misinformation} {Interventions}},
	url = {https://workshop-proceedings.icwsm.org/abstract.php?id=2025_10},
	doi = {10.36190/2025.10},
	language = {en-US},
	urldate = {2025-07-24},
	journal = {Workshop Proceedings of the 19th International AAAI Conference on Web and Social Media},
	author = {King, Catherine and Carragher, Peter and Carley, Kathleen M.},
	month = jun,
	year = {2025},
	pages = {10},
}

@inproceedings{ng2022online,
  title={Online coordination: methods and comparative case studies of coordinated groups across four events in the united states},
  author={Ng, Lynnette Hui Xian and Carley, Kathleen M},
  booktitle={Proceedings of the 14th ACM Web Science Conference 2022},
  pages={12--21},
  year={2022}
}

@article{carley2020social,
  title={Social cybersecurity: an emerging science},
  author={Carley, Kathleen M},
  journal={Computational and Mathematical Organization Theory},
  volume={26},
  number={4},
  pages={365--381},
  year={2020},
  publisher={Springer}
}

@article{bodeRelatedNewsThat2015,
	title = {In {Related} {News}, {That} {Was} {Wrong}: {The} {Correction} of {Misinformation} {Through} {Related} {Stories} {Functionality} in {Social} {Media}},
	volume = {65},
	issn = {0021-9916},
	shorttitle = {In {Related} {News}, {That} was {Wrong}},
	doi = {10.1111/jcom.12166},
	number = {4},
	urldate = {2024-05-15},
	journal = {Journal of Communication},
	author = {Bode, Leticia and Vraga, Emily K.},
	month = aug,
	year = {2015},
	pages = {619--638},
}

@article{smithCorrectingMisinformationNeuroscience2019,
	title = {Correcting {Misinformation} {About} {Neuroscience} via {Social} {Media}},
	volume = {41},
	issn = {1075-5470},
	url = {https://doi.org/10.1177/1075547019890073},
	doi = {10.1177/1075547019890073},
	language = {en},
	number = {6},
	urldate = {2024-05-15},
	journal = {Science Communication},
	author = {Smith, Ciarra N. and Seitz, Holli H.},
	month = dec,
	year = {2019},
	pages = {790--819},
}

@techreport{yadavPlatformInterventions2021,
	title = {Platform {Interventions}: {How} {Social} {Media} {Counters} {Influence} {Operations}},
	url = {https://carnegieendowment.org/2021/01/25/platform-interventions-how-social-media-counters-influence-operations-pub-83698},
	urldate = {2021-09-16},
	institution = {Carnegie Endowment for International Peace},
	author = {Yadav, Kamya},
	month = jan,
	year = {2021},
}

@article{pennycookShiftingAttentionAccuracy2021,
	title = {Shifting attention to accuracy can reduce misinformation online},
	copyright = {2021 The Author(s), under exclusive licence to Springer Nature Limited},
	issn = {1476-4687},
	url = {https://www.nature.com/articles/s41586-021-03344-2},
	doi = {10.1038/s41586-021-03344-2},
	language = {en},
	urldate = {2021-04-15},
	journal = {Nature},
	author = {Pennycook, Gordon and Epstein, Ziv and Mosleh, Mohsen and Arechar, Antonio A. and Eckles, Dean and Rand, David G.},
	month = mar,
	year = {2021},
	pages = {590-–595},
    volume = {592}
}

@article{epsteinDevelopingAccuracypromptToolkit2021,
	title = {Developing an accuracy-prompt toolkit to reduce {COVID}-19 misinformation online},
	url = {https://misinforeview.hks.harvard.edu/article/developing-an-accuracy-prompt-toolkit-to-reduce-covid-19-misinformation-online/},
	doi = {10.37016/mr-2020-71},
	language = {en-US},
	urldate = {2023-02-28},
	journal = {Harvard Kennedy School Misinformation Review},
	author = {Epstein, Ziv and Berinsky, Adam J. and Cole, Rocky and Gully, Andrew and Pennycook, Gordon and Rand, David G.},
	month = may,
	year = {2021},
    volume = {2},
    number = {3},
}

@article{bagoFakeNewsFast2020,
	title = {Fake news, fast and slow: {Deliberation} reduces belief in false (but not true) news headlines},
	volume = {149},
	issn = {1939-2222},
	shorttitle = {Fake news, fast and slow},
	doi = {10.1037/xge0000729},
	number = {8},
	journal = {Journal of Experimental Psychology: General},
	author = {Bago, Bence and Rand, David G. and Pennycook, Gordon},
	year = {2020},
	pages = {1608--1613},
}

@inproceedings{katsarosReconsideringTweetsIntervening2022,
	series = {{ICWSM}},
	title = {Reconsidering {Tweets}: {Intervening} during {Tweet} {Creation} {Decreases} {Offensive} {Content}},
	volume = {16},
	copyright = {Copyright (c) 2022 Association for the Advancement of Artificial Intelligence},
	shorttitle = {Reconsidering {Tweets}},
	url = {https://ojs.aaai.org/index.php/ICWSM/article/view/19308},
	doi = {10.1609/icwsm.v16i1.19308},
	language = {en},
	urldate = {2024-06-02},
	booktitle = {Proceedings of the {International} {AAAI} {Conference} on {Web} and {Social} {Media}},
	publisher = {Association for the Advancement of Artificial Intelligence},
	author = {Katsaros, Matthew and Yang, Kathy and Fratamico, Lauren},
	month = may,
	year = {2022},
	pages = {477--487},
}

@misc{porterfieldTwitterBeginsAsking2020,
	title = {Twitter {Begins} {Asking} {Users} {To} {Actually} {Read} {Articles} {Before} {Sharing} {Them}},
	url = {https://www.forbes.com/sites/carlieporterfield/2020/06/10/twitter-begins-asking-users-to-actually-read-articles-before-sharing-them/},
	language = {en},
	urldate = {2024-11-23},
	journal = {Forbes},
	author = {Porterfield, Carlie},
	month = jun,
	year = {2020},
}

@article{gillespieNotRecommendReduction2022,
	title = {Do {Not} {Recommend}? {Reduction} as a {Form} of {Content} {Moderation}},
	volume = {8},
	issn = {2056-3051},
	shorttitle = {Do {Not} {Recommend}?},
	url = {https://doi.org/10.1177/20563051221117552},
	doi = {10.1177/20563051221117552},
	number = {3},
	urldate = {2023-05-03},
	journal = {Social Media + Society},
	author = {Gillespie, Tarleton},
	month = jul,
	year = {2022},
	pages = {20563051221117552},
}

@article{kirchnerCounteringFakeNews2020,
	title = {Countering {Fake} {News}: {A} {Comparison} of {Possible} {Solutions} {Regarding} {User} {Acceptance} and {Effectiveness}},
	volume = {4},
	issn = {2573-0142},
	shorttitle = {Countering {Fake} {News}},
	url = {https://dl.acm.org/doi/10.1145/3415211},
	doi = {10.1145/3415211},
	language = {en},
	number = {CSCW2},
	urldate = {2023-05-01},
	journal = {Proceedings of the ACM on Human-Computer Interaction},
	author = {Kirchner, Jan and Reuter, Christian},
	month = oct,
	year = {2020},
	pages = {1--27},
}

@article{sharevskiMisinformationWarningsTwitters2022,
	title = {Misinformation warnings: {Twitter}’s soft moderation effects on {COVID}-19 vaccine belief echoes},
	volume = {114},
	issn = {0167-4048},
	shorttitle = {Misinformation warnings},
	url = {https://www.ncbi.nlm.nih.gov/pmc/articles/PMC8675217/},
	doi = {10.1016/j.cose.2021.102577},
	urldate = {2024-06-01},
	journal = {Computers \& Security},
	author = {Sharevski, Filipo and Alsaadi, Raniem and Jachim, Peter and Pieroni, Emma},
	month = mar,
	year = {2022},
	pmid = {34934255},
	pmcid = {PMC8675217},
	pages = {102577},
}

@techreport{barrettTacklingDomesticDisinformation2019,
	title = {Tackling {Domestic} {Disinformation}: {What} the {Social} {Media} {Companies} {Need} to {Do}},
	shorttitle = {Tackling {Domestic} {Disinformation}},
	url = {https://issuu.com/nyusterncenterforbusinessandhumanri/docs/nyu_domestic_disinformation_digital},
	language = {en},
	urldate = {2024-11-19},
	institution = {NYU Stern Center for Business and Human Rights},
	author = {Barrett, Paul},
	month = mar,
	year = {2019},
}

@article{kimPromotingOnlineCivility2022,
	title = {Promoting {Online} {Civility} {Through} {Platform} {Architecture}},
	volume = {1},
	copyright = {Copyright (c) 2022 Journal of Online Trust and Safety},
	issn = {2770-3142},
	url = {https://tsjournal.org/index.php/jots/article/view/54},
	doi = {10.54501/jots.v1i4.54},
	language = {en},
	number = {4},
	urldate = {2024-06-02},
	journal = {Journal of Online Trust and Safety},
	author = {Kim, Jisu and McDonald, Curtis and Meosky, Paul and Katsaros, Matthew and Tyler, Tom},
	month = sep,
	year = {2022},
}

@techreport{helmusRecommendationsCountering2021,
	title = {A {Compendium} of {Recommendations} for {Countering} {Russian} and {Other} {State}-{Sponsored} {Propaganda}},
	url = {https://www.rand.org/pubs/research_reports/RRA894-1.html},
	language = {en},
	urldate = {2021-06-08},
	institution = {RAND Corporation},
	author = {Helmus, Todd C. and Kepe, Marta},
	month = jun,
	year = {2021},
}

@misc{porterWhatsAppSaysIts2020,
	title = {{WhatsApp} says its forwarding limits have cut the spread of viral messages by 70 percent},
	url = {https://www.theverge.com/2020/4/27/21238082/whatsapp-forward-message-limits-viral-misinformation-decline},
	language = {en-US},
	urldate = {2025-03-05},
	journal = {The Verge},
	author = {Porter, Jon},
	month = apr,
	year = {2020},
}

@inproceedings{schaffnerCommunityGuidelinesMake2024,
	title = {``{Community} {Guidelines} {Make} this the {Best} {Party} on the {Internet}": {An} {In}-{Depth} {Study} of {Online} {Platforms}' {Content} {Moderation} {Policies}},
	shorttitle = {"{Community} {Guidelines} {Make} this the {Best} {Party} on the {Internet}"},
	url = {https://doi.org/10.1145/3613904.3642333},
	doi = {10.1145/3613904.3642333},
	urldate = {2025-01-03},
	booktitle = {Proceedings of the {CHI} {Conference} on {Human} {Factors} in {Computing} {Systems}},
	author = {Schaffner, Brennan and Bhagoji, Arjun Nitin and Cheng, Siyuan and Mei, Jacqueline and Shen, Jay L. and Wang, Grace and Chetty, Marshini and Feamster, Nick and Lakier, Genevieve and Tan, Chenhao},
	month = may,
	year = {2024},
	articleno = {486}, 
    numpages = {16},
    isbn = {9798400703300}, 
    publisher = {Association for Computing Machinery},
    series = {CHI '24}
}

@article{walterHowUnringBell2018,
	title = {How to unring the bell: {A} meta-analytic approach to correction of misinformation},
	volume = {85},
	issn = {0363-7751},
	shorttitle = {How to unring the bell},
	doi = {10.1080/03637751.2018.1467564},
	number = {3},
	urldate = {2023-05-01},
	journal = {Communication Monographs},
	author = {Walter, Nathan and Murphy, Sheila T.},
	month = jul,
	year = {2018},
	pages = {423--441},
}

@inproceedings{luEffectsAIbasedCredibility2022,
	series = {{CSCW}},
	title = {The {Effects} of {AI}-based {Credibility} {Indicators} on the {Detection} and {Spread} of {Misinformation} under {Social} {Influence}},
	volume = {6},
	url = {https://dl.acm.org/doi/10.1145/3555562},
	doi = {10.1145/3555562},
	urldate = {2024-06-01},
	booktitle = {Proceedings of the {ACM} on {Human}-{Computer} {Interaction}},
	publisher = {Association for Computing Machinery},
	author = {Lu, Zhuoran and Li, Patrick and Wang, Weilong and Yin, Ming},
	month = nov,
	year = {2022},
	pages = {461:1--461:27},
}

@article{khanFakeNewsOutbreak2021,
	title = {Fake news outbreak 2021: {Can} we stop the viral spread?},
	volume = {190},
	issn = {1084-8045},
	shorttitle = {Fake news outbreak 2021},
	url = {https://www.sciencedirect.com/science/article/pii/S1084804521001326},
	doi = {10.1016/j.jnca.2021.103112},
	language = {en},
	urldate = {2022-10-12},
	journal = {Journal of Network and Computer Applications},
	author = {Khan, Tanveer and Michalas, Antonis and Akhunzada, Adnan},
	month = sep,
	year = {2021},
	pages = {103112},
}

@misc{chiouFakeNewsAdvertising2018,
	type = {Working {Paper}},
	series = {Working {Paper} {Series}},
	title = {Fake {News} and {Advertising} on {Social} {Media}: {A} {Study} of the {Anti}-{Vaccination} {Movement}},
	shorttitle = {Fake {News} and {Advertising} on {Social} {Media}},
	url = {https://www.nber.org/papers/w25223},
	doi = {10.3386/w25223},
	urldate = {2023-05-03},
	publisher = {National Bureau of Economic Research},
	author = {Chiou, Lesley and Tucker, Catherine},
	month = nov,
	year = {2018},
}

@misc{borFactcheckingVideosReduce2020,
	title = {``{Fact}-checking" videos reduce belief in misinformation and improve the quality of news shared on {Twitter}},
	url = {https://osf.io/a7huq},
	doi = {10.31234/osf.io/a7huq},
	language = {en-us},
	urldate = {2024-05-15},
	publisher = {OSF},
	author = {Bor, Alexander and Osmundsen, Mathias and Rasmussen, Stig Hebbelstrup Rye and Bechmann, Anja and Petersen, Michael Bang},
	month = sep,
	year = {2020},
}

@misc{lewandowskyDebunkingHandbook2020,
	title = {Debunking {Handbook} 2020},
	copyright = {Databrary Access Agreement},
	url = {https://sks.to/db2020},
	doi = {10.17910/b7.1182},
	language = {en},
	urldate = {2023-09-11},
	publisher = {Databrary},
	author = {Lewandowsky, Stephan and Cook, John and Ecker, Ullrich and Albarracín, Dolores and Amazeen, Michelle A. and Kendeou, Panayiota and Lombardi, Doug and Newman, Eryn J. and Pennycook, Gordon and Porter, Ethan and Rand, David G. and Rapp, David N. and Reifler, Jason and Roozenbeek, Jon and Schmid, Philipp and Seifert, Colleen M. and Sinatra, Gale M. and Swire-Thompson, Briony and van der Linden, Sander and Vraga, Emily K. and Wood, Thomas J. and Zaragoza, Maria S.},
	year = {2020},
}

@article{chanDebunkingMetaAnalysisPsychological2017,
	title = {Debunking: {A} {Meta}-{Analysis} of the {Psychological} {Efficacy} of {Messages} {Countering} {Misinformation}},
	volume = {28},
	issn = {0956-7976},
	shorttitle = {Debunking},
	doi = {10.1177/0956797617714579},
	language = {en},
	number = {11},
	urldate = {2023-05-01},
	journal = {Psychological Science},
	author = {Chan, Man-pui Sally and Jones, Christopher R. and Hall Jamieson, Kathleen and Albarracín, Dolores},
	month = nov,
	year = {2017},
	pages = {1531--1546},
}

@article{bodeSeeSomethingSay2018,
	title = {See {Something}, {Say} {Something}: {Correction} of {Global} {Health} {Misinformation} on {Social} {Media}},
	volume = {33},
	issn = {1041-0236},
	shorttitle = {See {Something}, {Say} {Something}},
	doi = {10.1080/10410236.2017.1331312},
	number = {9},
	urldate = {2023-05-01},
	journal = {Health Communication},
	author = {Bode, Leticia and Vraga, Emily K.},
	month = sep,
	year = {2018},
	pages = {1131--1140},
}

@article{gorwaAlgorithmicContentModeration2020,
	title = {Algorithmic content moderation: {Technical} and political challenges in the automation of platform governance},
	volume = {7},
	issn = {2053-9517},
	shorttitle = {Algorithmic content moderation},
	url = {https://doi.org/10.1177/2053951719897945},
	doi = {10.1177/2053951719897945},
	number = {1},
	urldate = {2023-05-03},
	journal = {Big Data \& Society},
	author = {Gorwa, Robert and Binns, Reuben and Katzenbach, Christian},
	month = jan,
	year = {2020},
	pages = {2053951719897945},
}

@article{jhaverUsersWantPlatform2023,
	title = {Do users want platform moderation or individual control? {Examining} the role of third-person effects and free speech support in shaping moderation preferences},
	issn = {1461-4448},
	shorttitle = {Do users want platform moderation or individual control?},
	url = {https://doi.org/10.1177/14614448231217993},
	doi = {10.1177/14614448231217993},
	urldate = {2024-05-25},
	journal = {New Media \& Society},
	author = {Jhaver, Shagun and Zhang, Amy X.},
	month = dec,
	year = {2023},
    volume = {27},
    issue = {5},
}

@article{rauchfleischImpactDeplatformingFar2024,
	title = {The impact of deplatforming the far right: an analysis of {YouTube} and {BitChute}},
	volume = {27},
	issn = {1369-118X, 1468-4462},
	shorttitle = {The impact of deplatforming the far right},
	url = {https://www.tandfonline.com/doi/full/10.1080/1369118X.2024.2346524},
	doi = {10.1080/1369118X.2024.2346524},
	language = {en},
	number = {7},
	urldate = {2024-11-23},
	journal = {Information, Communication \& Society},
	author = {Rauchfleisch, Adrian and Kaiser, Jonas},
	month = may,
	year = {2024},
	pages = {1478--1496},
}

@article{thomasDisruptingHateEffect2023,
	title = {Disrupting hate: {The} effect of deplatforming hate organizations on their online audience},
	volume = {120},
	shorttitle = {Disrupting hate},
	url = {https://www.pnas.org/doi/10.1073/pnas.2214080120},
	doi = {10.1073/pnas.2214080120},
	number = {24},
	urldate = {2024-06-02},
	journal = {Proceedings of the National Academy of Sciences},
	author = {Thomas, Daniel Robert and Wahedi, Laila A.},
	month = jun,
	year = {2023},
	pages = {e2214080120},
}

@article{johnsLabellingShadowBans2024,
	title = {Labelling, shadow bans and community resistance: did {Meta}'s strategy to suppress rather than remove {COVID} misinformation and conspiracy theory on {Facebook} slow the spread?},
	issn = {1329-878X, 2200-467X},
	shorttitle = {Labelling, shadow bans and community resistance},
	url = {https://journals.sagepub.com/doi/10.1177/1329878X241236984},
	doi = {10.1177/1329878X241236984},
	language = {en},
	urldate = {2025-02-20},
	journal = {Media International Australia},
	author = {Johns, Amelia and Bailo, Francesco and Booth, Emily and Rizoiu, Marian-Andrei},
	month = mar,
	year = {2024},
    volume = {197},
    number = {1},
    pages = {222--241},
}

@article{westCensoredSuspendedShadowbanned2018,
	title = {Censored, suspended, shadowbanned: {User} interpretations of content moderation on social media platforms},
	volume = {20},
	issn = {1461-4448},
	shorttitle = {Censored, suspended, shadowbanned},
	url = {https://doi.org/10.1177/1461444818773059},
	doi = {10.1177/1461444818773059},
	number = {11},
	urldate = {2023-05-03},
	journal = {New Media \& Society},
	author = {West, Sarah Myers},
	month = nov,
	year = {2018},
	pages = {4366--4383},
}

@incollection{martinez-lopezSocialMediaMonetization2022,
	address = {Cham},
	title = {Social {Media} {Monetization} and {Demonetization}: {Risks}, {Challenges}, and {Potential} {Solutions}},
	isbn = {978-3-031-14575-9},
	shorttitle = {Social {Media} {Monetization} and {Demonetization}},
	url = {https://doi.org/10.1007/978-3-031-14575-9_13},
	language = {en},
	urldate = {2025-03-05},
	booktitle = {Social {Media} {Monetization}: {Platforms}, {Strategic} {Models} and {Critical} {Success} {Factors}},
	publisher = {Springer International Publishing},
	author = {Martínez-López, Francisco J. and Li, Yangchun and Young, Susan M.},
	editor = {Martínez-López, Francisco J. and Li, Yangchun and Young, Susan M.},
	year = {2022},
	doi = {10.1007/978-3-031-14575-9_13},
	pages = {185--214},
}

@article{morrowEmergingScienceContent2022,
	title = {The emerging science of content labeling: {Contextualizing} social media content moderation},
	volume = {73},
	issn = {2330-1643},
	shorttitle = {The emerging science of content labeling},
	url = {https://onlinelibrary.wiley.com/doi/abs/10.1002/asi.24637},
	doi = {10.1002/asi.24637},
	language = {en},
	number = {10},
	urldate = {2023-05-03},
	journal = {Journal of the Association for Information Science and Technology},
	author = {Morrow, Garrett and Swire-Thompson, Briony and Polny, Jessica Montgomery and Kopec, Matthew and Wihbey, John P.},
	year = {2022},
	pages = {1365--1386},
}

@article{allenScalingFactcheckingUsing2021,
	title = {Scaling up fact-checking using the wisdom of crowds},
	volume = {7},
	url = {https://www.science.org/doi/full/10.1126/sciadv.abf4393},
	doi = {10.1126/sciadv.abf4393},
	language = {en},
	number = {36},
	journal = {Science Advances},
	author = {Allen, Jennifer and Arechar, Antonio A. and Pennycook, Gordon and Rand, David G.},
	month = sep,
	year = {2021},
	pages = {eabf4393},
}

@inproceedings{allenBirdsFeatherDon2022,
	address = {New Orleans LA USA},
	title = {Birds of a feather don’t fact-check each other: {Partisanship} and the evaluation of news in {Twitter}’s {Birdwatch} crowdsourced fact-checking program},
	isbn = {978-1-4503-9157-3},
	shorttitle = {Birds of a feather don’t fact-check each other},
	url = {https://dl.acm.org/doi/10.1145/3491102.3502040},
	doi = {10.1145/3491102.3502040},
	language = {en},
	urldate = {2023-02-02},
	booktitle = {{CHI} {Conference} on {Human} {Factors} in {Computing} {Systems}},
	publisher = {ACM},
	author = {Allen, Jennifer and Martel, Cameron and Rand, David G.},
	month = apr,
	year = {2022},
	pages = {1--19},
}

@inproceedings{papakyriakopoulosImpactTwitterLabels2022,
	address = {New York, NY, USA},
	series = {{WWW} '22},
	title = {The {Impact} of {Twitter} {Labels} on {Misinformation} {Spread} and {User} {Engagement}: {Lessons} from {Trump}’s {Election} {Tweets}},
	isbn = {978-1-4503-9096-5},
	shorttitle = {The {Impact} of {Twitter} {Labels} on {Misinformation} {Spread} and {User} {Engagement}},
	url = {https://doi.org/10.1145/3485447.3512126},
	doi = {10.1145/3485447.3512126},
	urldate = {2022-09-28},
	booktitle = {Proceedings of the {ACM} {Web} {Conference} 2022},
	publisher = {Association for Computing Machinery},
	author = {Papakyriakopoulos, Orestis and Goodman, Ellen},
	month = apr,
	year = {2022},
	pages = {2541--2551},
}

@inproceedings{gaoLabelNotLabel2018,
	series = {{CSCW}},
	title = {To {Label} or {Not} to {Label}: {The} {Effect} of {Stance} and {Credibility} {Labels} on {Readers}' {Selection} and {Perception} of {News} {Articles}},
	volume = {2},
	shorttitle = {To {Label} or {Not} to {Label}},
	url = {https://dl.acm.org/doi/10.1145/3274324},
	doi = {10.1145/3274324},
	urldate = {2024-05-23},
	booktitle = {Proceedings of the {ACM} on {Human}-{Computer} {Interaction}},
    publisher = {Association for Computing Machinery},
	author = {Gao, Mingkun and Xiao, Ziang and Karahalios, Karrie and Fu, Wai-Tat},
	month = nov,
	year = {2018},
	pages = {55:1--55:16},
}

@article{nassettaStateMediaWarning2020,
	title = {State media warning labels can counteract the effects of foreign disinformation},
	volume = {1},
	url = {https://misinforeview.hks.harvard.edu/article/state-media-warning-labels-can-counteract-the-effects-of-foreign-misinformation/},
	doi = {10.37016/mr-2020-45},
	language = {en-US},
	number = {Special Issue on US Elections and Disinformation},
	urldate = {2024-05-15},
	journal = {Harvard Kennedy School Misinformation Review},
	author = {Nassetta, Jack and Gross, Kimberly},
	month = oct,
	year = {2020},
}

@article{niklewiczWeedingOutFake2017,
	title = {Weeding {Out} {Fake} {News}: {An} {Approach} to {Social} {Media} {Regulation}},
	volume = {16},
	issn = {1781-6858, 1865-5831},
	shorttitle = {Weeding {Out} {Fake} {News}},
	doi = {10.1007/s12290-017-0468-0},
	language = {en},
	number = {2},
	urldate = {2023-05-08},
	journal = {European View},
	author = {Niklewicz, Konrad},
	month = dec,
	year = {2017},
	pages = {335--335},
}

@article{tandocDiffusionDisinformationHow2020,
	title = {Diffusion of disinformation: {How} social media users respond to fake news and why},
	volume = {21},
	issn = {1464-8849},
	shorttitle = {Diffusion of disinformation},
	doi = {10.1177/1464884919868325},
	language = {en},
	number = {3},
	urldate = {2022-10-12},
	journal = {Journalism},
	author = {Tandoc, Edson C. and Lim, Darren and Ling, Rich},
	month = mar,
	year = {2020},
	pages = {381--398},
}

@article{badrinathanDonThinkThat2023,
	title = {“{I} {Don}’t {Think} {That}’s {True}, {Bro}!” {Social} {Corrections} of {Misinformation} in {India}},
	volume = {29},
	issn = {1940-1612},
	doi = {10.1177/19401612231158770},
	language = {en},
	urldate = {2023-03-09},
	journal = {The International Journal of Press/Politics},
	author = {Badrinathan, Sumitra and Chauchard, Simon},
	month = feb,
	year = {2023},
	pages = {394--416},
}

@misc{venturaMisinformationExposureTraditional2023,
	address = {Rochester, NY},
	type = {{SSRN} {Scholarly} {Paper}},
	title = {Misinformation {Exposure} {Beyond} {Traditional} {Feeds}: {Evidence} from a {WhatsApp} {Deactivation} {Experiment} in {Brazil}},
	shorttitle = {Misinformation {Exposure} {Beyond} {Traditional} {Feeds}},
	url = {https://papers.ssrn.com/abstract=4457400},
	doi = {10.2139/ssrn.4457400},
	language = {en},
	urldate = {2024-06-02},
	author = {Ventura, Tiago and Majumdar, Rajeshwari and Nagler, Jonathan and Tucker, Joshua A.},
	month = may,
	year = {2023},
}

@article{eckerCombiningRefutationsSocial2023,
	title = {Combining refutations and social norms increases belief change},
	volume = {76},
	issn = {1747-0226},
	url = {https://journals.sagepub.com/doi/10.1177/17470218221111750},
	doi = {10.1177/17470218221111750},
	language = {eng},
	number = {6},
	journal = {Quarterly Journal of Experimental Psychology (2006)},
	author = {Ecker, Ullrich K. H. and Sanderson, Jasmyne A. and McIlhiney, Paul and Rowsell, Jessica J. and Quekett, Hayley L. and Brown, Gordon DA and Lewandowsky, Stephan},
	month = jun,
	year = {2023},
	pmid = {35748514},
	pmcid = {PMC7614554},
	pages = {1275--1297},
}

@article{gimpelEffectivenessSocialNorms2021,
	title = {The {Effectiveness} of {Social} {Norms} in {Fighting} {Fake} {News} on {Social} {Media}},
	volume = {38},
	copyright = {© 2021 Taylor \& Francis Group, LLC},
	issn = {0742-1222},
	doi = {10.1080/07421222.2021.1870389},
	language = {EN},
	number = {1},
	urldate = {2024-06-02},
	journal = {Journal of Management Information Systems},
	author = {Gimpel, Henner and Heger, Sebastian and Olenberger, Christian and Utz, Lena},
	month = jan,
	year = {2021},
	pages = {196--221},
}

@article{avramExposureSocialEngagement2020,
	title = {Exposure to social engagement metrics increases vulnerability to misinformation},
	url = {https://misinforeview.hks.harvard.edu/article/exposure-to-social-engagement-metrics-increases-vulnerability-to-misinformation/},
	doi = {10.37016/mr-2020-033},
	language = {en-US},
	urldate = {2025-01-17},
	journal = {Harvard Kennedy School Misinformation Review},
	author = {Avram, Mihai and Micallef, Nicholas and Patil, Sameer and Menczer, Filippo},
	month = jul,
	year = {2020},
    volume = {1},
    number = {5},
}

@inproceedings{arifCloserLookSelfCorrecting2017,
	address = {New York, NY, USA},
	series = {{CSCW}},
	title = {A {Closer} {Look} at the {Self}-{Correcting} {Crowd}: {Examining} {Corrections} in {Online} {Rumors}},
	isbn = {978-1-4503-4335-0},
	shorttitle = {A {Closer} {Look} at the {Self}-{Correcting} {Crowd}},
	url = {https://dl.acm.org/doi/10.1145/2998181.2998294},
	doi = {10.1145/2998181.2998294},
	urldate = {2024-05-22},
	booktitle = {Proceedings of the 2017 {ACM} {Conference} on {Computer} {Supported} {Cooperative} {Work} and {Social} {Computing}},
	publisher = {Association for Computing Machinery},
	author = {Arif, Ahmer and Robinson, John J. and Stanek, Stephanie A. and Fichet, Elodie S. and Townsend, Paul and Worku, Zena and Starbird, Kate},
	month = feb,
	year = {2017},
	pages = {155--168},
}

@article{eckerCanYouBelieve2021,
	title = {Can you believe it? {An} investigation into the impact of retraction source credibility on the continued influence effect},
	volume = {49},
	issn = {1532-5946},
	shorttitle = {Can you believe it?},
	url = {https://doi.org/10.3758/s13421-020-01129-y},
	doi = {10.3758/s13421-020-01129-y},
	language = {en},
	number = {4},
	urldate = {2024-05-15},
	journal = {Memory \& Cognition},
	author = {Ecker, Ullrich K. H. and Antonio, Luke M.},
	month = may,
	year = {2021},
	pages = {631--644},
}

@article{lewandowskyCounteringMisinformation2021,
	title = {Countering {Misinformation} and {Fake} {News} {Through} {Inoculation} and {Prebunking}},
	volume = {0},
	issn = {1046-3283},
	doi = {10.1080/10463283.2021.1876983},
	number = {0},
	urldate = {2021-09-16},
	journal = {European Review of Social Psychology},
	author = {Lewandowsky, Stephan and van der Linden, Sander},
	month = feb,
	year = {2021},
	pages = {1--38},
}

@article{guessDigitalMediaLiteracy2020,
	title = {A digital media literacy intervention increases discernment between mainstream and false news in the {United} {States} and {India}},
	volume = {117},
	doi = {10.1073/pnas.1920498117},
	number = {27},
	urldate = {2023-05-01},
	journal = {Proceedings of the National Academy of Sciences},
	author = {Guess, Andrew M. and Lerner, Michael and Lyons, Benjamin and Montgomery, Jacob M. and Nyhan, Brendan and Reifler, Jason and Sircar, Neelanjan},
	month = jul,
	year = {2020},
	pages = {15536--15545},
}

@article{jeongMediaLiteracyInterventions2012,
	title = {Media {Literacy} {Interventions}: {A} {Meta}-{Analytic} {Review}},
	volume = {62},
	issn = {0021-9916},
	shorttitle = {Media {Literacy} {Interventions}},
	doi = {10.1111/j.1460-2466.2012.01643.x},
	language = {eng},
	number = {3},
	journal = {The Journal of Communication},
	author = {Jeong, Se-Hoon and Cho, Hyunyi and Hwang, Yoori},
	month = jun,
	year = {2012},
	pmid = {22736807},
	pmcid = {PMC3377317},
	pages = {454--472},
}

@article{roozenbeekPrebunkingInterventionsBased2020,
	title = {Prebunking interventions based on “inoculation” theory can reduce susceptibility to misinformation across cultures},
	volume = {1},
	doi = {10.37016//mr-2020-008},
	language = {en-US},
	number = {2},
	urldate = {2022-09-28},
	journal = {Harvard Kennedy School Misinformation Review},
	author = {Roozenbeek, Jon and Linden, Sander van der and Nygren, Thomas},
	month = feb,
	year = {2020},
}

@article{modirrousta-galianGamifiedInoculationInterventions2023,
	title = {Gamified inoculation interventions do not improve discrimination between true and fake news: {Reanalyzing} existing research with receiver operating characteristic analysis},
	volume = {152},
	issn = {1939-2222},
	shorttitle = {Gamified inoculation interventions do not improve discrimination between true and fake news},
	doi = {10.1037/xge0001395},
	number = {9},
	journal = {Journal of Experimental Psychology: General},
	author = {Modirrousta-Galian, Ariana and Higham, Philip A.},
	year = {2023},
	pages = {2411--2437},
}

@article{jones-jangDoesMediaLiteracy2021,
	title = {Does {Media} {Literacy} {Help} {Identification} of {Fake} {News}? {Information} {Literacy} {Helps}, but {Other} {Literacies} {Don}’t},
	volume = {65},
	issn = {0002-7642},
	shorttitle = {Does {Media} {Literacy} {Help} {Identification} of {Fake} {News}?},
	url = {https://doi.org/10.1177/0002764219869406},
	doi = {10.1177/0002764219869406},
	language = {en},
	number = {2},
	urldate = {2024-05-15},
	journal = {American Behavioral Scientist},
	author = {Jones-Jang, S. Mo and Mortensen, Tara and Liu, Jingjing},
	month = feb,
	year = {2021},
	pages = {371--388},
}

@article{tullyDesigningTestingNews2020,
	title = {Designing and {Testing} {News} {Literacy} {Messages} for {Social} {Media}},
	volume = {23},
	issn = {1520-5436},
	url = {https://doi.org/10.1080/15205436.2019.1604970},
	doi = {10.1080/15205436.2019.1604970},
	number = {1},
	urldate = {2024-05-15},
	journal = {Mass Communication and Society},
	author = {Tully, Melissa and Vraga, Emily K. and Bode, Leticia},
	month = jan,
	year = {2020},
	pages = {22--46},
}

@techreport{bradshawRoadAheadMapping2021,
	title = {The {Road} {Ahead}: {Mapping} {Civil} {Society} {Responses} to {Disinformation}},
	url = {https://www.ned.org/wp-content/uploads/2021/01/The-Road-Ahead-Mapping-Civil-Society-Responses-to-Disinformation-Bradshaw-Neudert-Jan-2021-2.pdf},
	urldate = {2023-05-04},
	institution = {National Endowment for Democracy},
	author = {Bradshaw, Samantha and Neudert, Lisa-Maria},
	month = jan,
	year = {2021},
}

@article{toffSocialMediaKilling2021,
	title = {Is {Social} {Media} {Killing} {Local} {News}? {An} {Examination} of {Engagement} and {Ownership} {Patterns} in {U}.{S}. {Community} {News} on {Facebook}},
	volume = {0},
	issn = {2167-0811},
	shorttitle = {Is {Social} {Media} {Killing} {Local} {News}?},
	url = {https://doi.org/10.1080/21670811.2021.1977668},
	doi = {10.1080/21670811.2021.1977668},
	number = {0},
	urldate = {2023-05-03},
	journal = {Digital Journalism},
	author = {Toff, Benjamin and Mathews, Nick},
	month = oct,
	year = {2021},
	pages = {1--20},
}

@techreport{barbashinImprovingWesternStrategy2018,
	title = {Improving the {Western} {Strategy} to {Combat} {Kremlin} {Propaganda} and {Disinformation}},
	url = {https://www.atlanticcouncil.org/wp-content/uploads/2018/06/Improving_the_Western_Strategy.pdf},
	language = {en},
	urldate = {2024-11-19},
	institution = {Atlantic Council: Eurasia Center},
	author = {Barbashin, Anton},
	month = may,
	year = {2018},
}

@incollection{bishopEthicalChallengesPublishing2017,
	series = {Advances in {Research} {Ethics} and {Integrity}},
	title = {Ethical {Challenges} of {Publishing} and {Sharing} {Social} {Media} {Research} {Data}},
	volume = {2},
	isbn = {978-1-78714-486-6 978-1-78714-485-9},
	urldate = {2023-05-03},
	booktitle = {The {Ethics} of {Online} {Research}},
	publisher = {Emerald Publishing Limited},
	author = {Bishop, Libby and Gray, Daniel},
	editor = {Woodfield, Kandy},
	month = jan,
	year = {2017},
	note = {https://doi.org/10.1108/S2398-601820180000002007},
	pages = {159--187},
}

@misc{povichInternetAdsAre2021,
	title = {Internet {Ads} {Are} a {Popular} {Tax} {Target} for {Both} {Parties}},
	url = {https://pew.org/3psT1Rq},
	language = {en},
	urldate = {2021-12-10},
	journal = {Pew},
	author = {Povich, Elaine S.},
	month = jun,
	year = {2021},
}

@article{assenmacherBenchmarkingCrisisSocial2022,
	title = {Benchmarking {Crisis} in {Social} {Media} {Analytics}: {A} {Solution} for the {Data}-{Sharing} {Problem}},
	volume = {40},
	issn = {0894-4393},
	shorttitle = {Benchmarking {Crisis} in {Social} {Media} {Analytics}},
	url = {https://doi.org/10.1177/08944393211012268},
	doi = {10.1177/08944393211012268},
	number = {6},
	urldate = {2023-05-03},
	journal = {Social Science Computer Review},
	author = {Assenmacher, Dennis and Weber, Derek and Preuss, Mike and Calero Valdez, André and Bradshaw, Alison and Ross, Björn and Cresci, Stefano and Trautmann, Heike and Neumann, Frank and Grimme, Christian},
	month = dec,
	year = {2022},
	pages = {1496--1522},
}

@techreport{wardleInformationDisorderInterdisciplinary2017,
	title = {Information {Disorder}: {Toward} an interdisciplinary framework for research and policy making},
	url = {https://rm.coe.int/information-disorder-toward-an-interdisciplinary-framework-for-researc/168076277c},
	language = {en},
	urldate = {2023-08-02},
	institution = {Council of Europe},
	author = {Wardle, Claire and Derakhshan, Hossein},
	month = sep,
	year = {2017},
}

@techreport{yadavCounteringInfluenceOperations2020,
	title = {Countering {Influence} {Operations}: {A} {Review} of {Policy} {Proposals} {Since} 2016},
	shorttitle = {Countering {Influence} {Operations}},
	url = {https://carnegieendowment.org/2020/11/30/countering-influence-operations-review-of-policy-proposals-since-2016-pub-83333},
	language = {en},
	urldate = {2023-05-01},
	institution = {Carnegie Endowment for International Peace},
	author = {Yadav, Kamya},
	month = nov,
	year = {2020},
}

@article{costelloDurablyReducingConspiracy2024,
	title = {Durably reducing conspiracy beliefs through dialogues with {AI}},
	volume = {385},
	url = {https://www.science.org/doi/10.1126/science.adq1814},
	doi = {10.1126/science.adq1814},
	number = {6714},
	urldate = {2024-10-04},
	journal = {Science},
	author = {Costello, Thomas H. and Pennycook, Gordon and Rand, David G.},
	month = sep,
	year = {2024},
	pages = {eadq1814},
}

@techreport{helmusGenerativeArtificialIntelligence2024,
	title = {Generative {Artificial} {Intelligence} {Threats} to {Information} {Integrity} and {Potential} {Policy} {Responses}},
	url = {https://www.rand.org/pubs/perspectives/PEA3089-1.html},
	language = {en},
	urldate = {2024-10-04},
	institution = {RAND Corporation},
	author = {Helmus, Todd C. and Chandra, Bilva},
	month = apr,
	year = {2024},
}

@article{bak-colemanCombiningInterventionsReduce2022,
	title = {Combining interventions to reduce the spread of viral misinformation},
	volume = {6},
	copyright = {2022 The Author(s)},
	issn = {2397-3374},
	doi = {10.1038/s41562-022-01388-6},
	language = {en},
	number = {10},
	urldate = {2022-09-26},
	journal = {Nature Human Behaviour},
	author = {Bak-Coleman, Joseph B. and Kennedy, Ian and Wack, Morgan and Beers, Andrew and Schafer, Joseph S. and Spiro, Emma S. and Starbird, Kate and West, Jevin D.},
	month = jun,
	year = {2022},
	pages = {1--9},
}

@article{myers2018censored,
  title={Censored, suspended, shadowbanned: User interpretations of content moderation on social media platforms},
  author={Myers West, Sarah},
  journal={New Media \& Society},
  volume={20},
  number={11},
  pages={4366--4383},
  year={2018},
  publisher={SAGE Publications Sage UK: London, England}
}

@incollection{ciampagliaDigitalMisinformationPipeline2018,
	address = {Wiesbaden},
	title = {The {Digital} {Misinformation} {Pipeline}},
	isbn = {978-3-658-19567-0},
	url = {https://doi.org/10.1007/978-3-658-19567-0_25},
	doi = {10.1007/978-3-658-19567-0_25},
	language = {en},
	urldate = {2025-04-17},
	booktitle = {Positive {Learning} in the {Age} of {Information}: {A} {Blessing} or a {Curse}?},
	publisher = {Springer Fachmedien},
	author = {Ciampaglia, Giovanni Luca},
	editor = {Zlatkin-Troitschanskaia, Olga and Wittum, Gabriel and Dengel, Andreas},
	year = {2018},
	pages = {413--421}
}

@article{gorwaWhatPlatformGovernance2019,
	title = {What is platform governance?},
	volume = {22},
	issn = {1369-118X},
	url = {https://doi.org/10.1080/1369118X.2019.1573914},
	doi = {10.1080/1369118X.2019.1573914},
	number = {6},
	urldate = {2025-03-07},
	journal = {Information, Communication \& Society},
	publisher = {Routledge},
	author = {Gorwa, Robert},
	month = may,
	year = {2019},
	pages = {854--871}
}

@article{liu2022implications,
	title = {Implications of {Revenue} {Models} and {Technology} for {Content} {Moderation} {Strategies}},
	volume = {41},
	issn = {0732-2399},
	url = {https://pubsonline.informs.org/doi/abs/10.1287/mksc.2022.1361},
	doi = {10.1287/mksc.2022.1361},
	number = {4},
	urldate = {2025-08-05},
	journal = {Marketing Science},
	publisher = {INFORMS},
	author = {Liu, Yi and Yildirim, Pinar and Zhang, Z. John},
	month = jul,
	year = {2022},
	pages = {831--847}
}

@techreport{mcclainHowUSPublic2025,
	title = {How the {U}.{S}. {Public} and {AI} {Experts} {View} {Artificial} {Intelligence}},
	url = {https://www.pewresearch.org/internet/2025/04/03/how-the-us-public-and-ai-experts-view-artificial-intelligence/},
	language = {en-US},
	urldate = {2025-05-02},
	institution = {Pew Research Center},
	author = {McClain, Colleen and Kennedy, Brian and Gottfried, Jeffrey and Anderson, Monica and Pasquini, Giancarlo},
	month = apr,
	year = {2025}
}

@article{aikinCorrectionMisleadingInformation2017,
	title = {Correction of misleading information in prescription drug television advertising: {The} roles of advertisement similarity and time delay},
	volume = {13},
	issn = {1551-7411},
	shorttitle = {Correction of misleading information in prescription drug television advertising},
	url = {https://www.sciencedirect.com/science/article/pii/S1551741116300146},
	doi = {10.1016/j.sapharm.2016.04.004},
	number = {2},
	urldate = {2024-05-15},
	journal = {Research in Social and Administrative Pharmacy},
	author = {Aikin, Kathryn J. and Southwell, Brian G. and Paquin, Ryan S. and Rupert, Douglas J. and O'Donoghue, Amie C. and Betts, Kevin R. and Lee, Philip K.},
	month = mar,
	year = {2017},
	pages = {378--388},
}

@article{murtagh2014ward,
  title={Ward’s hierarchical agglomerative clustering method: which algorithms implement Ward’s criterion?},
  author={Murtagh, Fionn and Legendre, Pierre},
  journal={Journal of Classification},
  volume={31},
  number={3},
  pages={274--295},
  year={2014},
  publisher={Springer}
}

@Manual{baseR2025,
    title = {R: A Language and Environment for Statistical Computing},
    author = {{R Core Team}},
    organization = {R Foundation for Statistical Computing},
    address = {Vienna, Austria},
    year = {2025},
    url = {https://www.R-project.org/},
  }


\appendix
\section{Intervention Categorization}
\label{sec:app_literature_interventions}

This appendix describes the literature of interventions, in which we used to develop our categorization taxonomy. We selected four prominent review articles about misinformation interventions, all published in different fields and journals (\citet{aghajariReviewingInterventionsAddress2023} in an HCI conference, \citet{blairInterventionsCounterMisinformation2024} in a psychology venue, \citet{courchesneReviewSocialScience2021} in an interdisciplinary social science journal, and \citet{kozyrevaToolboxIndividuallevelInterventions2024} in Nature Human Behavior). Table \ref{tab:lit-summary} summarizes the intervention categories analyzed by each paper in alphabetical order.

The \citet{courchesneReviewSocialScience2021} article classifies types of interventions based on those publicly announced as having been implemented by various social media platforms. These 10 categories are advertisement policy, content labeling, content / account moderation, content reporting, content distribution / sharing, disinformation disclosure, disinformation literacy, redirection, security / verification, and other. The article notes that fact-checking was included in the disinformation disclosure category, so this is categorized as fact-checking in Table \ref{tab:lit-summary}.

The \citet{aghajariReviewingInterventionsAddress2023} article categorizes interventions primarily based on the driver of the misinformation that each intervention targets: Content, Source, Individual Users, or Community. For content-based interventions, they discuss fact-checking, warning labels, and platform alterations (reducing the size of misinformation). For source-based interventions, they discuss source credibility labels and the crowdsourcing of those labels. For user-based strategies, they consider media literacy, accuracy prompts, and account removal. Lastly, for community-based interventions, they address social norms.

\begin{table}[htp]
    \centering
    \caption{Categorizations of misinformation interventions in the literature.}
    \begin{tabular}
    {>{\raggedright\arraybackslash}p{1.25in}>{\centering\arraybackslash}p{1in}
                >{\centering\arraybackslash}p{1in}
                >{\centering\arraybackslash}p{1in}
                >{\centering\arraybackslash}p{1in}}
      \toprule
      \textbf{Type}  & \textbf{Courchesne et al. (2021) \cite{courchesneReviewSocialScience2021}} &  \textbf{Aghajari et al. (2023) \cite{aghajariReviewingInterventionsAddress2023}}  & \textbf{Blair et al. (2024) \cite{blairInterventionsCounterMisinformation2024}} & \textbf{Kozyreva et al. (2024) \cite{kozyrevaToolboxIndividuallevelInterventions2024}}\\
      \midrule
      Account Moderation & $\times$ & & & \\
      Account Removal & & $\times$ & & \\
      Accuracy Prompts & & $\times$ & $\times$ & $\times$\\
      Advertising Policy & $\times$ & & & \\
      Content Distribution & $\times$ & & & \\
      Content Labeling & $\times$ & & & \\
      Content Moderation & $\times$ & & & \\
      Context Labels & & & $\times$ & \\
      Crowdsourcing & & $\times$ & & \\
      Debunking & & & $\times$ & $\times$\\
      Fact-Checking & $\times$ & $\times$ & & $\times$\\
      Friction & & & $\times$ & $\times$\\
      Inoculation & & & $\times$ & $\times$\\
      Journalist Training & & & $\times$ & \\
      Lateral Reading & & & & $\times$\\
      Media Literacy & $\times$ & $\times$ & $\times$ & $\times$\\
      Platform Alterations & & $\times$ & $\times$ & \\
      Politician Messaging & & & $\times$ & \\
      Redirection & $\times$ & & & \\
      Reporting & $\times$ & & & \\
      Security/Verification & $\times$ & & & \\
      Social Norms & & $\times$ & $\times$ & $\times$\\
      Source Credibility Labels & & $\times$ & $\times$ & $\times$\\
      Warning Labels & & $\times$ & & $\times$\\
      Other & $\times$ & &  &  \\
      \bottomrule
    \end{tabular}
    \label{tab:lit-summary}
\end{table}

The \citet{blairInterventionsCounterMisinformation2024} and \citet{kozyrevaToolboxIndividuallevelInterventions2024} articles
categorize interventions not just in terms of which parts of the platforms are affected, but also what part of the social media misinformation pipeline is targeted (such as the creation, spread, belief phases as well as prevention - see Figure \ref{fig:information_lifecycle}). More specifically, the \citet{blairInterventionsCounterMisinformation2024} article categorizes 11 types of interventions into four main groups: Informational (targets belief), Educational (targets prevention), Sociopsychological (targets spread), and Institutional (external). Information interventions include inoculation / prebunking, debunking, credibility labels / tags, and contextual labels / tags. These interventions aim to correct misinformation.  Under educational interventions, they define media literacy. This category of interventions seeks to prevent belief in misinformation. Sociopsychological interventions include accuracy prompts, friction, and social norms. These interventions aim to discourage users from spreading misinformation. Lastly, they define platform alterations, politician messaging, and journalist training under institutional interventions. 

The \citet{kozyrevaToolboxIndividuallevelInterventions2024}. article classifies nine types of interventions into three main categories \cite{kozyrevaToolboxIndividuallevelInterventions2024}: Nudges (targets spread), Boosts and Educational Interventions (targets prevention), and Refutation Strategies (targets belief). The nudges category includes  accuracy prompts, friction, and social norms. These interventions address sharing behavior by discouraging users from further distributing misinformation. The boosts/educational interventions category includes inoculation, lateral reading and verification strategies, and media literacy tips. These interventions aim to increase user competence in evaluating content online and discerning misinformation or untrustworthy content from reputable information.  Lastly, the refutation strategies category includes debunking and rebuttals, warning and fact-checking labels, and source credibility labels. 

Many of the review articles used a similar categorization of countermeasures; however, there is no common typology \cite{courchesneReviewSocialScience2021,aghajariReviewingInterventionsAddress2023,blairInterventionsCounterMisinformation2024,kozyrevaToolboxIndividuallevelInterventions2024}. Many of these defined categories overlap (e.g., lateral reading skills vs. media literacy) or are sub-categories of other categories. For example, redirection is a form of content distribution. Similarly, context labels and source credibility labels are types of content labeling. Additionally, specific categories are absent, particularly user-led or institutional interventions such as government regulation. User-based measures are an often overlooked aspect in the fight against misinformation.  Individual-level debunking, especially from trusted messengers, is effective in various contexts \cite{badrinathanDonThinkThat2023,bodeSeeSomethingSay2018,lewandowskyDebunkingHandbook2020}. 
In addition to user reporting and social norms, users can block others or engage in social corrections. When misinformation is successfully posted on social media, other users serve as the first line of defense since they can flag or debunk it. While social corrections can be considered a type of debunking or fact-checking, the social context may be important to distinguish.

By synthesizing the categorizations used by four review articles, we developed eight general categories of countermeasures, as shown in Table \ref{tab1}. These categories are primarily classified by the part of the information lifecycle targeted and the part of the platform affected. The following subsections describe these eight general categories in more detail, as well as the subcategories of intervention types often described in the literature for each general category.

\subsection{Account Moderation}\label{app-sec:am}
Account moderation refers to platform interventions directed at user accounts. These interventions include actions such as account suspensions or account removals, limiting account visibility through shadowbanning, or demonetizing accounts, which restricts their ability to monetize content\cite{courchesneReviewSocialScience2021, westCensoredSuspendedShadowbanned2018}. Specific interventions include:
\begin{itemize}
    \item \textbf{Account Suspension} - Account suspension involves temporarily suspending or permanently banning accounts that repeatedly violate platform policies, like repeatedly disseminating misinformation. Deplatforming describes a coordinated effort to remove a high-risk account across multiple platforms, rather than just on a single platform \cite{rauchfleischImpactDeplatformingFar2024, thomasDisruptingHateEffect2023}.
    \item \textbf{Shadow Banning} - Shadow banning suppresses an account's reach by limiting its visibility to others, while leaving the account itself still active. Users are often not notified that their content has been throttled \cite{johnsLabellingShadowBans2024, westCensoredSuspendedShadowbanned2018}.
    \item \textbf{Demonetization} - Demonetization refers to platforms removing or limiting an account's monetization privileges \cite{martinez-lopezSocialMediaMonetization2022}. Platforms typically apply such restrictions after repeated policy violations. 
\end{itemize}

\subsection{Content Moderation}\label{app-sec:cm}
Content moderation refers to a broad range of interventions regarding the way content is presented, distributed, or displayed on social media. These interventions range from evaluating the accuracy of information to modifying its visibility or circulation, and may be implemented through human moderators, users or automated systems \cite{jiangTradeoffcenteredFrameworkContent2023,schaffnerCommunityGuidelinesMake2024}. Specific interventions include:
\begin{itemize}
    \item \textbf{Misinformation Detection} - Misinformation detection refers to the use of computational methods to identify content that is potentially false, misleading, or otherwise unreliable. These systems typically serve as an upstream component of content moderation, identifying content that may be subsequently reviewed, labeled, downranked or removed \cite{khanFakeNewsOutbreak2021,luEffectsAIbasedCredibility2022}.
    \item \textbf{Fact-checking} - Fact-checking assesses the accuracy of a claim by comparing it against available evidence. It can be carried out by experts, journalists, and platforms, and includes both text and video-based formats \cite{borFactcheckingVideosReduce2020, walterHowUnringBell2018}.
    \item \textbf{Debunking} - Debunking goes beyond determining whether a claim is accurate by providing additional evidence, explanations or contextual information to correct misleading claims. In this sense, debunking can be understood as a form of 
    "narrative intervention" that identifies falsehoods and provides a more coherent and accurate account of the issue \cite{lewandowskyDebunkingHandbook2020,chanDebunkingMetaAnalysisPsychological2017}.
    \item \textbf{Algorithmic Moderation} - Algorithmic content moderation refers to the use of automated systems to identify, evaluate and manage content at scale. Such systems would automatically fact-check, label, remove, downrank or otherwise reduce the visibility of content that violate policies, like misinformation \cite{barrettTacklingDomesticDisinformation2019,bodeSeeSomethingSay2018,gillespieNotRecommendReduction2022,gorwaAlgorithmicContentModeration2020}. Algorithmic moderation can also function as a virality circuit breaker, temporarily limiting the amplification of rapidly spreading, unverified content, until its accuracy can be assessed \cite{DemocracyDesignSocial2024}.
    \item \textbf{User Control} - User control shifts some moderation responsibilities from the platforms to the users, allowing users to take a more active role in deciding what appears in their feeds, rather than than leaving that decision entirely to the platform's algorithms \cite{jhaverUsersWantPlatform2023}.
\end{itemize}

\subsection{Content Distribution}
\label{app-sec:cd}
Content distribution refers to a broad category of interventions that shape how content circulates, reaches audiences and is encountered on social media platforms. Rather than directly removing or modifying content, these interventions influence the pathways through which information is spread on social media.
Specific interventions include: 
\begin{itemize}
    \item \textbf{Account Limits} - Another type of content distribution is to impose account limits on users to restrict their forwarding capabilities, which caps the number of recipients the message can be forwarded to \cite{porterWhatsAppSaysIts2020}, resharing limits that constrain repeated dissemination, or disable sharing functionality after content has had several rounds of redistribution \cite{DemocracyDesignSocial2024}.
    \item \textbf{Friction} - Friction introduces additional steps or delays, such as an extra click or a confirmation window, into the process of sharing content so that users can pause and possibly reconsider before sharing \cite{bagoFakeNewsFast2020,katsarosReconsideringTweetsIntervening2022}. One example of such friction would be to prevent users from re-sharing content that they had not opened and prompting them to consider reading the article first instead \cite{porterfieldTwitterBeginsAsking2020}.
    \item \textbf{Accuracy Prompts} - Accuracy prompts, sometimes referred to as "nudges," are brief questions or reminders about the importance of accurate information sent to users and intended to bring accuracy to the front of their minds when a user is deciding whether or not to post a piece of content \cite{pennycookShiftingAttentionAccuracy2021}. One implementation involves asking users to evaluate the accuracy of one or more headlines to reflect the importance of sharing accurate news before they can continue using the platform \cite{epsteinDevelopingAccuracypromptToolkit2021}.
    \item \textbf{Redirection} - Redirection intercepts a search query on a potentially harmful or dangerous topic and directs the user toward vetted and official information. In some cases, it may suppress the search results entirely \cite{bodeRelatedNewsThat2015, smithCorrectingMisinformationNeuroscience2019}. Sometimes, platforms may limit access to the requested content altogether. A common example is surfacing guidelines from a public health organization, such as the CDC or the WHO, above or next to general search results when users search for a topic like COVID-19 vaccines. Similarly, in the immediate lead-up to an election in the United States, many platforms highlight how to find one’s polling place using official government sources. \cite{courchesneReviewSocialScience2021,yadavPlatformInterventions2021}.
    \item \textbf{Advertising Policy} - Advertising policies govern how advertisements can be created, distributed, targeted or displayed on social media platforms \cite{barrettTacklingDomesticDisinformation2019,courchesneReviewSocialScience2021,helmusRecommendationsCountering2021}. These policies govern the rules that platforms or regulators can impose on advertising content, and they may include restrictions on the type of content that can run and how it must be labeled or displayed \cite{barrettTacklingDomesticDisinformation2019,courchesneReviewSocialScience2021,helmusRecommendationsCountering2021}. Examples include banning political ads, prohibiting the use of AI or manipulated content in political or targeted ads, requiring ads to undergo a fact-checking process before posting, and labeling ads as "paid for" \cite{aikinCorrectionMisleadingInformation2017, chiouFakeNewsAdvertising2018,courchesneReviewSocialScience2021}.
    \item \textbf{Platform Alterations} - Platform alterations refer to interface- or architecture-level changes on social media platforms that shape how content is surfaced, sized, or displayed to users, which in turn affects how users interact with it \cite{kirchnerCounteringFakeNews2020}. Such interventions can reduce the size or visibility of a post \cite{gillespieNotRecommendReduction2022,kirchnerCounteringFakeNews2020,sharevskiMisinformationWarningsTwitters2022} or alter the platform interface to restructure how users can share content, such as directing posts toward more specific rather than general groups \cite{kimPromotingOnlineCivility2022}.
\end{itemize}

\subsection{Generative AI-specific}\label{app-sec:ai}
This category is for interventions that explicitly use generative AI tools like chatbots or LLM-written content to counter misinformation. Specific interventions include:
\begin{itemize}
    \item \textbf{Generative AI Content} - Using AI-generated content to create rebuttals against misinformation or to develop educational initiatives. 
    
    \item \textbf{Generative AI Chatbots} - Deploying conversational AI systems to dialogue with users as a way to reduce their belief in conspiracy theories or false claims \cite{costelloDurablyReducingConspiracy2024}. 
\end{itemize}
There are other generative AI-adjacent interventions, such as prohibiting the use of AI or manipulated content to produce deepfakes (Content Moderation), banning the use of AI in political or targeted advertising (Content Distribution), or requiring clear disclosure on any ads that incorporate AI-generated images, videos, or audio (Content Labeling) \cite{helmusGenerativeArtificialIntelligence2024}. However, these types of interventions already fit into pre-existing categories.

\subsection{Content Labeling}\label{app-sec:cl}
Content labeling refers to interventions that attach disclosures, warnings or supplementary information to content in order to communicate its accuracy, credibility, source, or broader context. These interventions provide users with information through labels that help them evaluate the content they encounter on social media. These labels include fact-checks (see Content Moderation section) or source information or credibility \cite{morrowEmergingScienceContent2022,yadavPlatformInterventions2021}. Specific interventions include:
\begin{itemize}
    \item \textbf{Crowdsourcing} - Crowdsourcing involves drawing on evaluations from ordinary users to access information and provide labels or contextual information. It shifts the labeling task away from professionals and onto the user base, then aggregates user judgments into a label or note \cite{allenScalingFactcheckingUsing2021}. One prominent implementation of this approach is X's Community Notes program, in which users collaboratively contribute and evaluate contextual notes that attached to the tweets \cite{allenBirdsFeatherDon2022}.
    \item \textbf{Context Labels} - Context labels attach additional contextual information to posts so that users can better interpret or evaluate the content. These labels provide relevant background, clarification or evidence to the post. X's Community Notes are context labels, and the program uses crowdsourcing to create and maintain them \cite{allenBirdsFeatherDon2022}.
    \item \textbf{Source Labels} - Source credibility labels provide users with information about the reliability, identity, or institutional affiliation of the source behind a post. These interventions may, for example, communicate assessments of a news source's reliability \cite{gaoLabelNotLabel2018} or identify accounts associated with government officials or state-run media sources \cite{nassettaStateMediaWarning2020}.
    \item \textbf{Warning Labels} - Warning labels flag posts based on the post's content or sources that may be false, misleading, or otherwise warrant additional scrutiny \cite{papakyriakopoulosImpactTwitterLabels2022}. Some implementations require an active click to bypass an interstitial screen before the user can view or share the underlying content \cite{sharevskiMisinformationWarningsTwitters2022}.
    \item \textbf{Notification} - Notifications directly inform users when content they have been posted, shared or otherwise interacted with has subsequently been flagged to contain harmful or false information, or originating from state-run media sources \cite{courchesneReviewSocialScience2021}.
\end{itemize}

\subsection{User-based Measures}\label{app-sec:user}
User-based measures refer to interventions that empower individuals and communities to respond directly to misinformation and shape their own information environments \cite{tandocDiffusionDisinformationHow2020}.  Specific interventions include:
\begin{itemize}
    \item \textbf{Social Norms} - Social norm interventions leverage community expectations and social signals to encourage more responsible information-sharing behaviors. Platforms can support these norms by designing features or metrics that reward accuracy rather than engagement \cite{eckerCombiningRefutationsSocial2023,gimpelEffectivenessSocialNorms2021,avramExposureSocialEngagement2020}.
    \item \textbf{Social Corrections} - Social corrections are peer-to-peer corrections in which one user directly fact-checks another’s post, sometimes occurring in the open (via a reply) and other times occurring privately (via a direct message) \cite{badrinathanDonThinkThat2023,bodeSeeSomethingSay2018}.
    
    \item \textbf{Retractions} - Retractions occur when users or organizations delete or correct a false post that was previously published. It may also involve efforts to notify users who previously viewed the post. \cite{arifCloserLookSelfCorrecting2017,eckerCanYouBelieve2021}.
    \item \textbf{User Reporting} - Reporting enables users to flag potentially harmful or misleading posts or accounts for platform review \cite{niklewiczWeedingOutFake2017,zhouEffectiveReportingSystem2024}.
    \item \textbf{Blocking} - Blocking gives users a way to reduce their exposure to unwanted content, such as content from specific users or on specific topics \cite{tandocDiffusionDisinformationHow2020}. 
    
    \item \textbf{Other/User Empowerment} - Individualized measures provide users with tools or interventions that encourage self-moderation and behavioral change, such as nudges which will encourage users to reduce their social media use \cite{venturaMisinformationExposureTraditional2023}.
\end{itemize}

\subsection{Media Literacy and Education}\label{app-sec:ml}
Media literacy and education encompasses interventions designed to strengthen individuals' ability to critically evaluate and engage with information.  These approaches seek to improve the public’s ability to navigate digital sources and to think critically about the media they encounter \cite{guessDigitalMediaLiteracy2020,jeongMediaLiteracyInterventions2012}. Specific interventions include:
\begin{itemize}
    \item \textbf{Media Literacy} - Media literacy and related educational interventions include a range of initiatives intended to develop or assess media literacy skills. These efforts may provide practical guidance for identifying misinformation \cite{guessDigitalMediaLiteracy2020}, evaluate individuals' information, digital, or news literacy \cite{jones-jangDoesMediaLiteracy2021}, or develop and test communication strategies for improving the effectiveness of media literacy education \cite{tullyDesigningTestingNews2020}.
    \item \textbf{Inoculation} - Inoculation, sometimes called "pre-bunking" in the literature, aims to build resistance to false information before exposure occurs. These interventions typically warn individuals about common misleading arguments or manipulation techniques, and explain how they operate, making users better prepared to recognize and resist subsequent misinformation \cite{lewandowskyCounteringMisinformation2021}. 
    \item \textbf{Games} - Games like fake news games are a specific type of educational tool that use interactive, gamified experiences to teach users how to recognize misinformation and strengthen their critical evaluation skills. These games often expose players to common misinformation strategies in a controlled environment, helping them identify similar techniques in future encounters \cite{maertensLongtermEffectivenessInoculation2021,modirrousta-galianGamifiedInoculationInterventions2023,roozenbeekPrebunkingInterventionsBased2020}.
\end{itemize}

\subsection{Institutional Measures}\label{app-sec:im}
Institutional measures refer to interventions undertaken by governments, civil society organizations, news media, and other institutions \cite{bradshawRoadAheadMapping2021}. Specific interventions include:
\begin{itemize}
    \item \textbf{Government Regulation} - Government regulation encompasses laws and regulations designed to govern platforms and increase their accountability. These measures can be implemented at the local, state, or federal level \cite{niklewiczWeedingOutFake2017,rochefortRegulatingSocialMedia2020,yadavCounteringInfluenceOperations2020}. Examples include modifying Section 230, developing comprehensive privacy laws similar to Europe's GDPR, taking anti-trust action to break up monopolistic technology companies \cite{rochefortRegulatingSocialMedia2020}, or restriction of micro-targeted advertising  \cite{povichInternetAdsAre2021}.
    \item \textbf{Media Support} - Media support involves efforts to strengthen the broader news ecosystem, including funding local media outlets, giving credible local reporting greater visibility on social media platforms \cite{toffSocialMediaKilling2021}, and providing training and resources to support high-quality reporting \cite{barbashinImprovingWesternStrategy2018,bradshawRoadAheadMapping2021}.    
    \item \textbf{Data Sharing} - Data sharing involves social media platforms sharing raw data and internal findings with researchers, thereby supporting greater transparency and independent investigation \cite{assenmacherBenchmarkingCrisisSocial2022,bishopEthicalChallengesPublishing2017,barrettTacklingDomesticDisinformation2019}.
    \item \textbf{Other Institutional Measures} - Other institutional measures include developing resources and tools to support civil society and strengthening coordination and collaboration among governments, researchers, media organizations, platforms and other relevant institutions \cite{bradshawRoadAheadMapping2021,wardleInformationDisorderInterdisciplinary2017}.
\end{itemize}

\section{Operationalized Interventions}
\label{sec:app_detailed_interventions}

This section describes the 40 operationalized interventions participants were asked about in the expert survey. Each intervention is a specific implementation of one or more interventions described in the literature and in the previous appendix. On the survey, participants were given the definitions of the specific intervention implementations, as shown in the first column of the following table.

\clearpage
\onecolumn

\small
\begin{longtable}{p{0.72\linewidth} p{0.235\linewidth}}
\toprule
\textbf{Specified Intervention Implementations} & \textbf{Intervention Type(s)} \\
\midrule
\endfirsthead

\toprule
\textbf{Specified Intervention Implementations} & \textbf{Intervention Type(s)} \\
\midrule
\endhead

\midrule
\multicolumn{2}{r}{\emph{Continued on next page}} \\
\endfoot

\bottomrule
\caption{Operationalized misinformation interventions across the eight general categories.}
\label{tab:operationalized-interventions}
\endlastfoot

\multicolumn{2}{l}{\textbf{Account Moderation}}\\
\addlinespace[2pt]
\textit{Account Suspensions} - Permanently or temporarily ban certain users. & Account Suspension \\
\textit{Deplatforming} - Coordinated efforts to ban prominent misinformation spreaders. & Account Suspension \\
\textit{Shadow Banning} - Limit the visibility and the spread of posts from certain policy-violating accounts. & Shadow Banning \\
\textit{Demonetization} - Remove or restrict monetization features for a user account. & Demonetization \\

\addlinespace
\multicolumn{2}{l}{\textbf{Content Moderation}}\\
\addlinespace[2pt]
\textit{Content Removal} - Remove posts verified to contain misinformation. & Misinformation Detection \\
\textit{Ban Deepfakes} - Prohibit the use of AI to misrepresent the speech or actions of public figures. & Misinformation Detection \\
\textit{Downranking} - De-emphasize posts in news feeds that contain misinformation. & Algorithmic Moderation \\
\textit{Upranking} - Uprank high-quality news sources and downrank low-quality ones. & Algorithmic Moderation \\
\textit{Virality Circuit Breakers} - Briefly halt algorithmic amplification of unverified content.& Algorithmic Moderation \\ 
\textit{User Control} - Give users more control over the algorithms powering their news feeds. & User Control \\

\addlinespace
\multicolumn{2}{l}{\textbf{Content Distribution}}\\
\addlinespace[2pt]
\textit{Limit Forwarding} - Cap the number of users one can forward a given message to. & Account Limits \\
\textit{Limit Resharing} - Remove share buttons on posts after several levels of sharing. & Account Limits \\
\textit{Friction} - Temporarily delay users from posting content they did not open via a pop-up. & Friction \\
\textit{Accuracy Prompts} - Nudge or remind people to consider accuracy before posting or sharing content. & Accuracy Prompts \\
\textit{Redirection} - Redirect users to other content, such as official content or no content. & Redirection \\
\textit{Ban Political Ads} - Ban political ads on social media platforms. & Advertising Policy \\
\textit{Fact-Check Ads} - Fact-check ads on social media platforms. & Advertising Policy \\
\textit{AI in Ads} - Prohibit the use of AI or manipulated content in political or targeted ads. &  Advertising Policy \\
\textit{Platform Alterations} - Reduce the size or visibility of a post containing misinformation. & Platform Alterations \\

\addlinespace
\multicolumn{2}{l}{\textbf{Generative AI}}\\
\addlinespace[2pt]
\textit{Gen AI Content} - Generate rebuttals to misinformation or create educational initiatives. & Generative AI \\
\textit{Gen AI Chatbots} - Use chatbots to reduce belief in conspiracy theories or misinformation. & Generative AI \\

\addlinespace
\multicolumn{2}{l}{\textbf{Content Labeling}}\\
\addlinespace[2pt]
\textit{Crowdsourcing} - Labels generated by the public rather than professionals (e.g., Community Notes). & Crowdsourcing, Context Labels \\
\textit{Fact Check Labels} - Link information from third-party fact-checkers on misinformation posts. & Fact-Checking, Context Labels \\
\textit{Source Labels} - Labels indicating the reliability of news sources. & Source Labels \\
\textit{Government Labels} - Label the accounts of government officials or state-run media. & Source Labels \\
\textit{AI Disclosure} - Require disclosures on advertisements that use AI-generated content. & Source Labels, Context Labels\\
\textit{Click-through Warning Labels} - Place misinformation posts behind click-through labels containing facts and context. & Warning Labels \\

\addlinespace
\multicolumn{2}{l}{\textbf{User-Based Measures}}\\
\addlinespace[2pt]
\textit{Social Norms} - Use social or community norms to encourage social corrections and self-corrections. & Social Corrections, Retractions \\
\textit{Alter Platform Metrics} - Reward accuracy rather than engagement to discourage misinformation sharing. & Social Norms\\
\textit{Reporting} - Improved reporting functionality and transparency. & User Reporting\\

\addlinespace
\multicolumn{2}{l}{\textbf{Media Literacy / Education}}\\
\addlinespace[2pt]
\textit{Digital Media Literacy} - Invest in and promote content on how to critically evaluate online information. & Media Literacy \\
\textit{Inoculation} - Inoculate people against misinformation with games or videos. & Inoculation, Games \\

\addlinespace
\multicolumn{2}{l}{\textbf{Institutional Measures}}\\
\addlinespace[2pt]
\textit{Government Regulation} - Hold companies accountable for content on their platforms. & Government Regulation \\
\textit{Privacy Legislation} - Develop comprehensive privacy legislation, similar to Europe's GDPR. & Government Regulation \\
\textit{Anti-Trust Action} - Break up monopolistic big tech companies. & Government Regulation \\
\textit{Taxes / Fines} - Tax or fine social media companies for their use of personal user data. & Government Regulation \\
\textit{Targeted Advertising} - Limit or ban micro-targeted advertising. & Government Regulation \\
\textit{Media Support} - Promote and invest in local media. & Media Support \\
\textit{Journalism Support} - Support and train the next generation of journalists. & Media Support \\
\textit{Data Sharing} - Have companies regularly release data and internal research reports to researchers. & Data Sharing \\
\end{longtable}

\normalsize

\clearpage
\twocolumn

\end{document}